%% file: main.tex
\documentclass{article} 
\usepackage{iclr2026_conference,times}

\input{math_commands.tex}

\usepackage{hyperref}
\usepackage{url}

\usepackage[utf8]{inputenc}
\usepackage{amsmath,amssymb}
\usepackage{booktabs}
\usepackage{hyperref}
\usepackage{xcolor}

\usepackage{graphicx}
\graphicspath{{images/}}

\newcommand{\cmark}{\ding{51}}
\newcommand{\xmark}{\ding{55}}
\usepackage{pifont}
\renewcommand{\cmark}{\ding{51}}
\renewcommand{\xmark}{\ding{55}}

\title{Open-Source Image Editing Models Are Zero-Shot Vision Learners}

\author{Wei Liu\thanks{Email: \texttt{wei.liu.llm@gmail.com}}, Jiaxin Lin, Rui Chen \\
Tencent Inc.\\
}

\iclrfinalcopy 
\begin{document}

\maketitle
\pagestyle{plain}
\thispagestyle{plain}

\input{tex/1.abstract}
\input{tex/2.introduction}
\input{tex/3.related_work}
\input{tex/4.method}

\input{tex/5.experiments}
\input{tex/6.conclusion.tex}


\bibliography{main.bbl}
\bibliographystyle{iclr2026_conference}


\end{document}

%% file: math_commands.tex
\usepackage{amsmath,amsfonts,bm}

\def\eqref#1{equation~\ref{#1}}

\def\1{\bm{1}}

\DeclareMathAlphabet{\mathsfit}{\encodingdefault}{\sfdefault}{m}{sl}
\SetMathAlphabet{\mathsfit}{bold}{\encodingdefault}{\sfdefault}{bx}{n}



%% file: tex/1.abstract.tex
\begin{abstract}
Recent studies have shown that large generative models can solve vision tasks they were not explicitly trained for. However, existing evidence relies on closed-source models~(Veo~3, Nano Banana Pro) or requires task-specific instruction tuning, leaving open whether publicly available image-editing models possess zero-shot vision abilities out of the box.

We conduct a systematic evaluation of three open-source image-editing models---Qwen-Image-Edit, FireRed-Image-Edit, and LongCat-Image-Edit---on dense visual prediction tasks \emph{without any fine-tuning}. We benchmark monocular depth estimation on NYUv2 and DIODE, surface normal estimation on NYUv2, and semantic segmentation on Cityscapes, covering both geometric and semantic scene understanding.

Results show that open-source image-editing models exhibit non-trivial zero-shot visual understanding. On NYUv2 surface normals, FireRed-Image-Edit achieves a mean angular error of $17.69^\circ$, surpassing the fine-tuned Marigold ($20.86^\circ$) and matching the instruction-tuned Vision Banana ($17.78^\circ$) without any task-specific training. On NYUv2 depth estimation, LongCat-Image-Edit obtains $\delta_1{=}0.822$ with affine alignment, and Qwen-Image-Edit leads on DIODE Indoor ($\delta_1{=}0.868$). On Cityscapes semantic segmentation, Qwen-Image-Edit reaches 25.7 mIoU at the 19-class level and 49.5 mIoU at a coarser 7-category level. By comparing three independently trained editors, we test whether zero-shot vision ability is an emergent property of image-editing pretraining rather than a model-specific artifact. Code, evaluation scripts, and all results are publicly released to serve as a reproducible baseline for future work.
\end{abstract}

%% file: tex/2.introduction.tex
\section{Introduction}

Image generation and editing models have evolved from tools for producing visually plausible images into systems that follow complex multimodal instructions~\citep{firered2025edit}. Given an input image and a text instruction, modern image editors can add objects, remove regions, modify attributes, and preserve scene layout while changing visual content. This instruction-following behavior suggests that such models may encode more than appearance priors: to edit an image coherently and reliably, a model must infer object boundaries, scene geometry, material layout, and spatial relationships.

This observation raises a natural question: \emph{can image-editing models serve as zero-shot vision learners?} Instead of training a specialized network for each visual prediction task, one can describe the desired output using language and ask an image editor to produce a visual representation of the answer. For example, monocular depth estimation can be prompted as generating a grayscale depth map, surface normal estimation as generating a standard normal map, and semantic segmentation as painting each class with a specified color. The generated image can then be decoded back into a dense prediction for quantitative evaluation~\citep{gabeur2026image}.

Recent work provides partial evidence for this idea, but important gaps remain.~\citet{wiedemer2025video} demonstrate that Veo~3, a large closed-source video model, can zero-shot solve segmentation, edge detection, and visual reasoning tasks. However, they do not evaluate geometric understanding tasks such as depth estimation or surface normal estimation, and the model is not publicly available.~\citet{zuo2025nano} evaluate Nano Banana Pro on 14 low-level vision tasks including dehazing, super-resolution, and denoising, but do not test any geometric or semantic scene understanding tasks.~\citet{gabeur2026image} show that an image generator can be instruction-tuned into a generalist vision model (Vision Banana) that handles depth, surface normals, and segmentation, but they report only fine-tuned results and do not provide a zero-shot quantitative baseline. Across all three studies, the models are closed-source or custom-trained, and no study compares multiple independently trained models to test whether zero-shot vision behavior is model-specific or shared across the image-editing paradigm.

In this paper, we fill these gaps by conducting a systematic zero-shot evaluation of \emph{open-source} image-editing models on dense visual prediction tasks. We focus on three representative task families:
\begin{itemize}
    \item \textbf{Monocular depth estimation}, which requires understanding relative and metric scene geometry. We evaluate on NYUv2~\citep{silberman2012nyu} and DIODE Indoor/Outdoor~\citep{vasiljevic2019diode}.
    \item \textbf{Surface normal estimation}, which probes local 3D orientation and geometric structure. We evaluate on NYUv2~\citep{silberman2012nyu}.
    \item \textbf{Semantic segmentation}, which tests whether image editors can follow class-to-color instructions and localize semantic categories. We evaluate on Cityscapes~\citep{cordts2016cityscapes}, focusing on color-guided mask generation.
\end{itemize}
We omit instance segmentation from the quantitative evaluation because current zero-shot image editing models lack the output-format consistency needed to reliably decode individual instances.

We evaluate three publicly available image-editing models---Qwen-Image-Edit~\citep{qwen2025edit}, FireRed-Image-Edit~\citep{firered2025edit}, and LongCat-Image-Edit~\citep{longcat2025edit}---all in a zero-shot setting, without task-specific training or prompt tuning. Each task is expressed through text prompts, and generated images are decoded into quantitative predictions using deterministic, task-specific visual codecs.

Our results reveal non-trivial zero-shot geometric and semantic ability. On NYUv2 surface normals, FireRed-Image-Edit achieves a mean angular error of $17.69^\circ$, surpassing the fine-tuned Marigold ($20.86^\circ$) and matching the instruction-tuned Vision Banana ($17.78^\circ$) without any task-specific training. For depth estimation, LongCat-Image-Edit captures relative scene geometry on NYUv2 ($\delta_1{=}0.822$) and Qwen-Image-Edit leads on DIODE Indoor ($\delta_1{=}0.868$), both with affine alignment. For semantic segmentation on Cityscapes, Qwen-Image-Edit reaches 25.7 mIoU at the 19-class level and 49.5 mIoU at 7 coarse categories, indicating meaningful scene-level understanding. These results also reveal a key bottleneck: quantitative dense prediction depends not only on visual understanding, but also on whether the editor follows the requested output format precisely. This motivates a broader comparison across independently trained image editors to test whether such visual understanding is an emergent property of image-editing pretraining.

Our contributions are as follows:
\begin{itemize}
    \item We provide the first systematic zero-shot evaluation of open-source image-editing models as dense vision learners, covering three independently trained models across depth, surface normal estimation, and semantic segmentation tasks.
    \item We demonstrate that simple task prompts and lightweight deterministic decoders can extract dense predictions from zero-shot editors without model fine-tuning or instruction tuning.
    \item We show that zero-shot image editors recover meaningful relative geometry and surface orientation, with FireRed-Image-Edit matching the instruction-tuned Vision Banana result on NYUv2 normals.
    \item We release\footnote{\url{https://github.com/liuwei1206/zeroshot-VL}} all code, evaluation scripts, and results to serve as a public baseline for future work on generative models as general-purpose vision systems.
\end{itemize}

%% file: tex/3.related_work.tex
\section{Related Work}

\paragraph{Dense Prediction in Computer Vision.}
Depth estimation, surface normal estimation, and semantic segmentation are core dense prediction tasks. Traditional approaches train task-specific models with supervised annotations on benchmarks such as NYUv2~\citep{silberman2012nyu}, DIODE~\citep{vasiljevic2019diode}, Cityscapes~\citep{cordts2016cityscapes}, and ADE20K~\citep{zhou2017ade20k}. More recent methods such as Marigold~\citep{ke2024marigold} and Lotus~\citep{he2024lotus} leverage diffusion priors for monocular depth and surface normal estimation, achieving strong results after task-specific adaptation. These methods show that generative priors are useful for dense prediction, but they remain specialized vision models rather than zero-shot image editors.

\paragraph{Generative Models as Zero-Shot Vision Learners.}
A growing line of work investigates whether generative models possess zero-shot vision abilities without task-specific training.~\citet{wiedemer2025video} demonstrate that Veo~3 can solve a broad range of visual tasks, including segmentation, edge detection, keypoint localization, and visual reasoning. However, their study does not evaluate geometric understanding tasks such as depth estimation or surface normal estimation, and Veo~3 is closed-source.

\citet{zuo2025nano} evaluate Nano Banana Pro, a closed-source image generation model, on low-level restoration and enhancement tasks such as dehazing, super-resolution, denoising, and deraining. Their results provide evidence that generative models can perform visual tasks beyond image synthesis, but they do not cover geometric scene understanding or semantic segmentation.

\citet{gabeur2026image} propose Vision Banana, which instruction-tunes an image generator to handle depth estimation, surface normal estimation, and segmentation. Their results show that image generation can serve as a unified interface for vision tasks after output-format alignment. However, they do not report zero-shot quantitative results for the released base model, leaving open how much dense visual understanding exists before instruction tuning.

\paragraph{Gap Addressed by This Work.}
Table~\ref{tab:related_comparison} summarizes the positioning of our work relative to prior studies. Existing evidence relies on closed-source systems or custom-trained models, and no prior work provides a zero-shot quantitative evaluation of geometric understanding across multiple independently trained public image editors. Our goal is not to train a new generalist model, but to measure what released image-editing models can already do without adaptation.

\begin{table}[t]
\centering
\small
\begin{tabular}{lcccc}
\hline
 & Public & Zero- & Geometry & Cross- \\
 & model & shot & eval. & model \\
\hline
Wiedemer et al. & \xmark & \cmark & \xmark & \xmark \\
Zuo et al. & \xmark & \cmark & \xmark & \xmark \\
Vision Banana & \xmark & \xmark & \cmark & \xmark \\
\textbf{Ours} & \cmark & \cmark & \cmark & \cmark \\
\hline
\end{tabular}
\caption{Comparison with prior work. ``Public model'' indicates publicly available model weights or checkpoints. ``Geometry eval.'' indicates evaluation on depth and/or surface normals. ``Cross-model'' indicates comparison of multiple independently trained models.}
\label{tab:related_comparison}
\end{table}

\paragraph{Image Editing Models.}
Instruction-guided image editing models transform an input image according to a natural-language prompt. Unlike text-to-image models, image editors must preserve relevant input content while modifying selected attributes or regions. We study Qwen-Image-Edit~\citep{qwen2025edit}, FireRed-Image-Edit~\citep{firered2025edit}, and LongCat-Image-Edit~\citep{longcat2025edit}, three publicly available editing checkpoints released by different groups. Using them as a cross-model testbed allows us to ask whether zero-shot dense prediction is tied to the image-editing paradigm rather than to a single model or training recipe.

%% file: tex/4.method.tex
\section{Method}
\label{sec:method}

We evaluate image-editing models as zero-shot dense vision learners using a unified prompt-and-decode pipeline, following the output-as-image formulation of Vision Banana~\citep{gabeur2026image}. Given an input image $I$ and a task prompt $p_t$, an image-editing model $G$ produces an RGB output image $\hat{Y}_{\mathrm{rgb}} = G(I, p_t)$. A deterministic task decoder $D_t$ maps this image to the task prediction $\hat{Y}$:
\begin{equation}
    \hat{Y} = D_t(G(I, p_t)).
\end{equation}
No model parameters are updated; the only task-specific components are the prompts and deterministic decoders.

\subsection{Zero-Shot Protocol}

Each image-editing model is used in its released form. We do not perform supervised fine-tuning or LoRA adaptation. For each task, the same prompt template and decoder are applied to all models. Thus, if a model fails to follow the requested output format, the decoded prediction degrades directly and the error is reflected in the metrics.

\subsection{Depth Estimation}

For monocular depth estimation, the prompt asks the model to convert the input image into a \textbf{grayscale depth map}, where luminance monotonically encodes distance from the camera (brighter = nearer, darker = farther). Unlike Vision Banana~\citep{gabeur2026image}, which uses a bijective false-color depth encoding with instruction-tuned models, we request grayscale depth output directly. This simpler format avoids reliance on a specific colormap and is more suitable for zero-shot editors that may not reproduce complex color sequences reliably.

We decode the generated grayscale depth map into a scalar relative-depth prediction by extracting perceived luminance with ITU-R BT.709~\citep{itu2015bt709} weights:
\begin{equation}
    \hat{d}_i = 0.2126\, R_i + 0.7152\, G_i + 0.0722\, B_i,
\end{equation}
where $(R_i, G_i, B_i)$ are the normalized RGB values at pixel $i$. This luminance-based decoder is robust to slightly tinted grayscale outputs and preserves the predicted depth ordering.

Because zero-shot image editors are not trained to produce calibrated metric depth, raw evaluation would conflate geometric structure with global scale and offset errors. We therefore use \textbf{affine alignment}: for each image, we fit a scale $a$ and offset $b$ on valid pixels via least squares, yielding $\hat{d}_{\mathrm{aligned}} = a \hat{d} + b$. This evaluates relative scene geometry independently of metric calibration.

\subsection{Surface Normal Estimation}

For surface normal estimation, the prompt asks the model to generate a standard normal map. Following the parameterization used in Vision Banana~\citep{gabeur2026image}, each unit normal vector $\mathbf{n}=(n_x, n_y, n_z)$ is encoded as:
\begin{equation}
    \mathrm{RGB} = \frac{\mathbf{n} + 1}{2}.
\end{equation}
The generated RGB image is decoded by applying the inverse mapping and normalizing each vector to unit length.

\paragraph{Automatic Axis Convention Calibration.}
Surface normal conventions can differ across generated outputs and dataset annotations. Rather than manually specifying an axis mapping for each model--dataset pair, we calibrate it automatically. We enumerate all 48 possible conventions (6 axis permutations $\times$ 8 sign-flip combinations) and evaluate each on a small calibration subset of 5 samples. The convention with the lowest mean angular error is applied to the full evaluation set. This calibration is performed once per model--dataset pair and cached.

This step does not tune the image-editing model or decoder parameters; it only resolves a discrete coordinate-convention ambiguity. Such alignment is standard when comparing systems that may use different 3D coordinate conventions.

\subsection{Semantic Segmentation}

For semantic segmentation, we use color-specified mask generation in the spirit of Vision Banana~\citep{gabeur2026image}. To isolate mask generation from class discovery, each prompt lists only the classes present in the ground-truth annotation and assigns them intuitive color names (e.g.\ ``Convert this photo into a color-coded map: the road red, the car white, \ldots and everything else black.''). This oracle class-list setting evaluates whether zero-shot editors can localize the specified categories and follow the requested palette.

The generated image is resized to the input resolution with nearest-neighbor interpolation. Each pixel is assigned to the class whose prompted color is nearest in RGB space:
\begin{equation}
    \hat{y}_{i} = \arg\min_{c \in \mathcal{C}_{\mathrm{img}}} \| \hat{r}_{i} - r_c \|_2,
\end{equation}
where $\mathcal{C}_{\mathrm{img}}$ is the set of classes mentioned in the prompt, $\hat{r}_{i}$ is the generated RGB value at pixel $i$, and $r_c$ is the target color for class $c$. A black background class absorbs pixels not assigned to any prompted foreground category.

In addition to the standard 19-class Cityscapes evaluation, we report a 7-category coarse evaluation that groups the classes into flat, construction, nature, sky, human, vehicle, and object. This coarser evaluation reduces fine-grained label confusion and highlights broad scene-layout understanding.

\subsection{Exclusion of Instance Segmentation}

We exclude instance segmentation from quantitative evaluation because it requires stronger output-format alignment than the other tasks. Unlike semantic segmentation, where colors are specified in the prompt, instance segmentation requires the model to assign a different color to each object instance while the number of instances is unknown. In zero-shot editor outputs, instance masks are often not piecewise-constant: colors may vary within an instance, boundaries may be blended, and background artifacts may appear. These artifacts make color clustering unreliable, leading to merged or fragmented instances.

Vision Banana addresses this issue through per-class inference and instruction tuning, which aligns the model to render each instance as a unique, solid-color region before color clustering~\citep{gabeur2026image}. Since our goal is to evaluate out-of-the-box image editors rather than instruction-tuned format following, we leave reliable zero-shot instance decoding to future work.

%% file: tex/5.experiments.tex
\section{Experiments}

\paragraph{Evaluated Models.}
We evaluate three publicly available image-editing models: Qwen-Image-Edit-2511~\citep{qwen2025edit}, FireRed-Image-Edit-1.1~\citep{firered2025edit}, and LongCat-Image-Edit~\citep{longcat2025edit}. All models are used zero-shot with released checkpoints and no task-specific adaptation. We use a fixed random seed throughout. For Qwen-Image-Edit and FireRed-Image-Edit, we use 40 diffusion steps with classifier-free guidance scale 4.0; for LongCat-Image-Edit, we use 50 diffusion steps with guidance scale 5.0.

\subsection{Depth Estimation}

\paragraph{Experimental Setup.}
Depth estimation tests whether an editor can infer 3D scene structure from a single RGB image. We evaluate on NYUv2~\citep{silberman2012nyu} (654 indoor validation images, 480$\times$640), DIODE Indoor~\citep{vasiljevic2019diode} (771 validation images, 768$\times$1024), and DIODE Outdoor (446 validation images, 768$\times$1024).

\paragraph{Implementation.}
Each model is prompted to generate a grayscale depth map, where pixel brightness encodes relative distance from the camera (e.g.\ ``Convert this image into a grayscale depth map with smooth gradual transitions. Nearby objects appear bright, distant objects appear dark.''). We resize the generated output to the input resolution with bilinear interpolation and decode it into depth by extracting perceived luminance (ITU-R BT.709 weights).

\paragraph{Metrics.}
We report $\delta_1$, AbsRel, and RMSE. Since zero-shot editors are not trained for metric calibration, we report affine-aligned metrics: a scale and offset are fitted per image before evaluation, isolating relative scene geometry from global calibration errors.

\begin{figure*}[t]
\centering
\includegraphics[width=0.245\textwidth,height=0.245\textheight,keepaspectratio]{depth_origin.png}
\hfill
\includegraphics[width=0.245\textwidth,height=0.245\textheight,keepaspectratio]{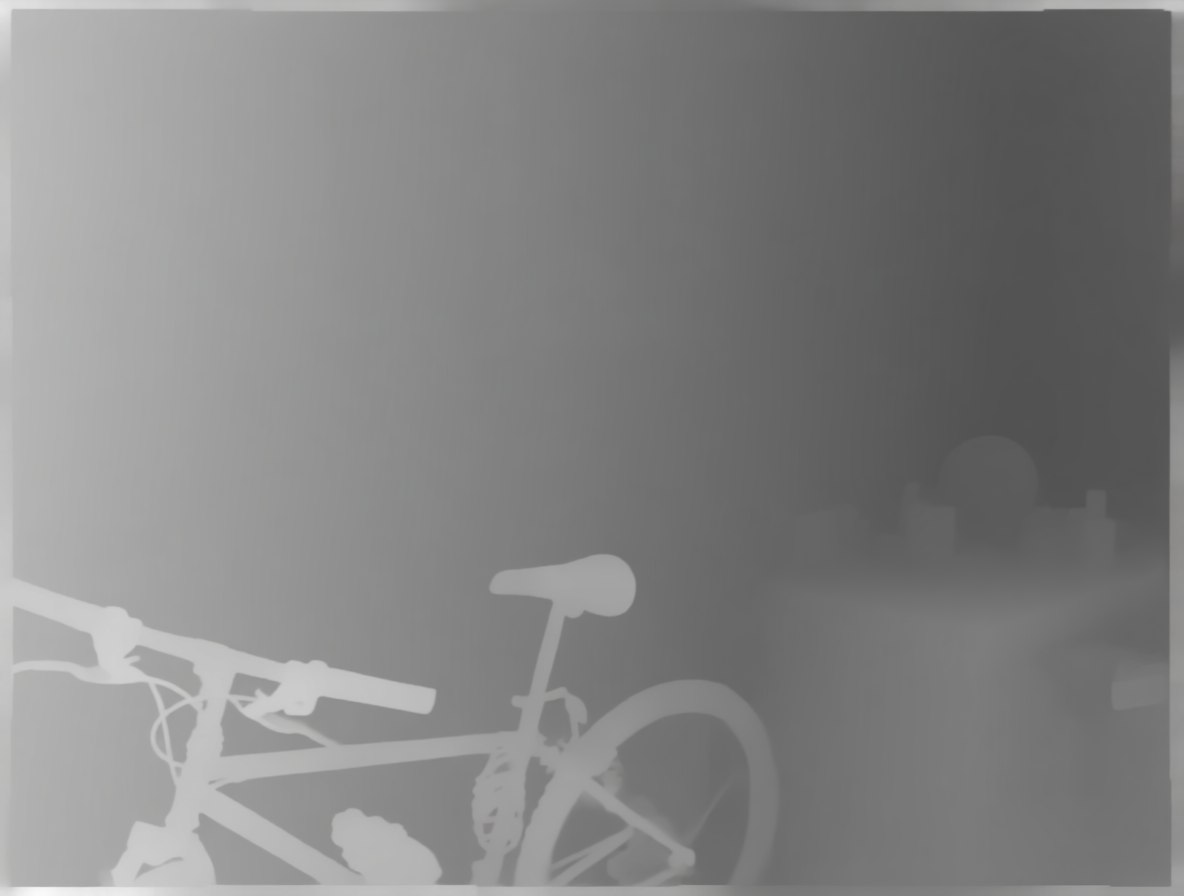}
\hfill
\includegraphics[width=0.245\textwidth,height=0.245\textheight,keepaspectratio]{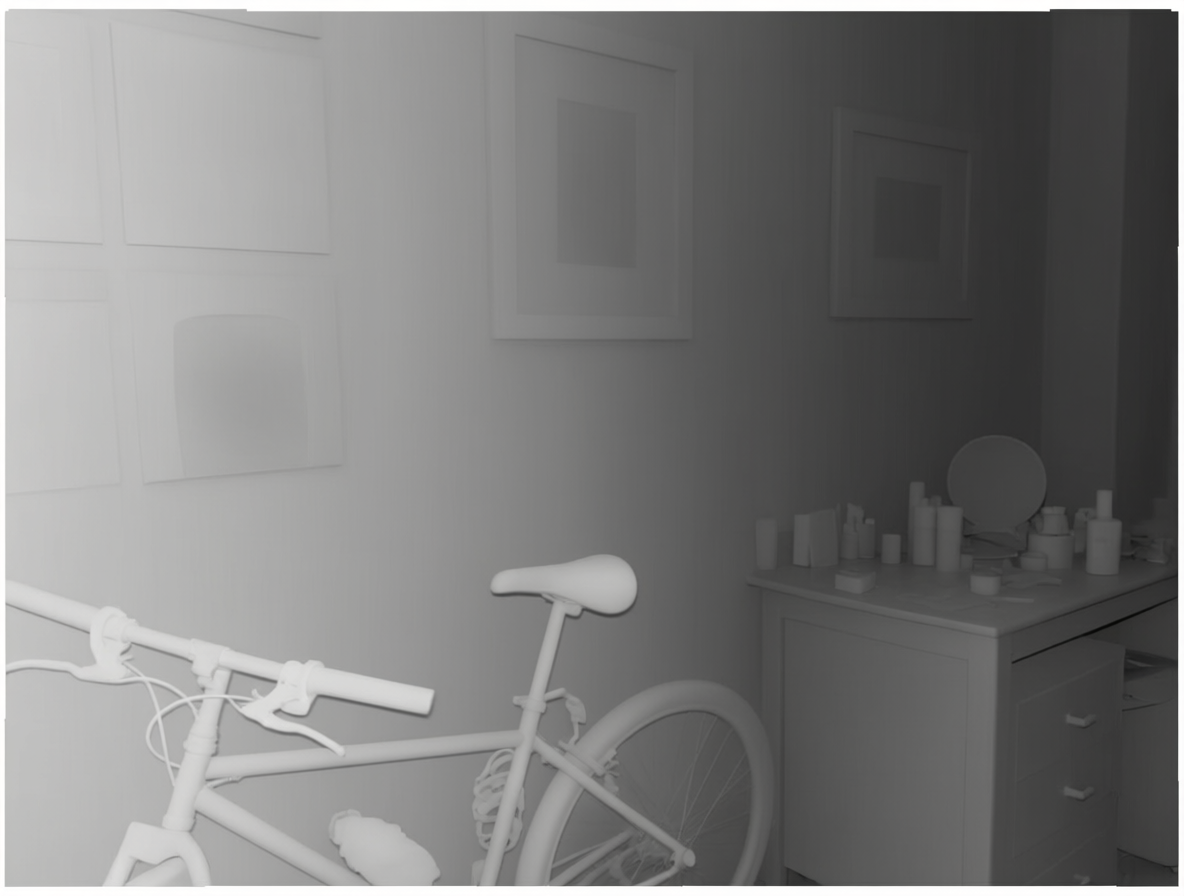}
\hfill
\includegraphics[width=0.245\textwidth,height=0.245\textheight,keepaspectratio]{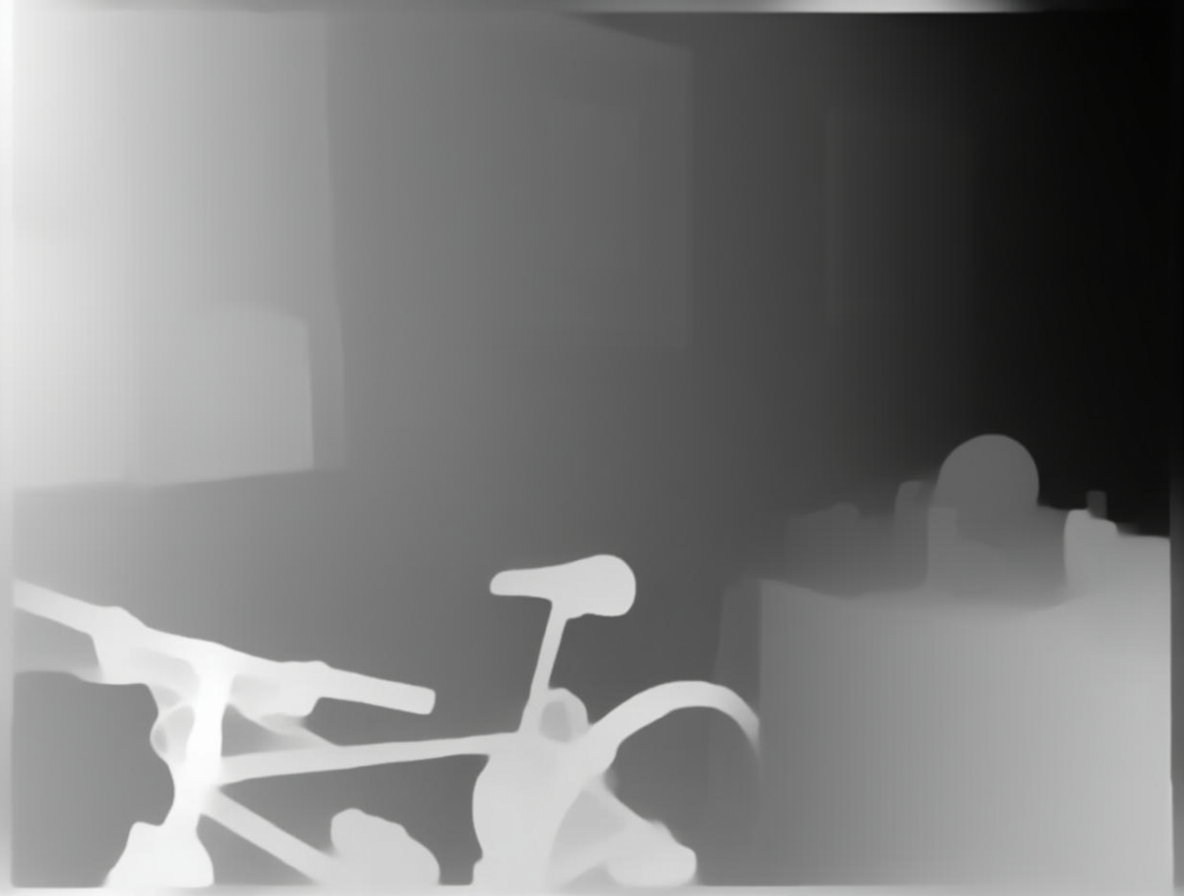}

\vspace{0.2em}

\includegraphics[width=0.245\textwidth,height=0.245\textheight,keepaspectratio]{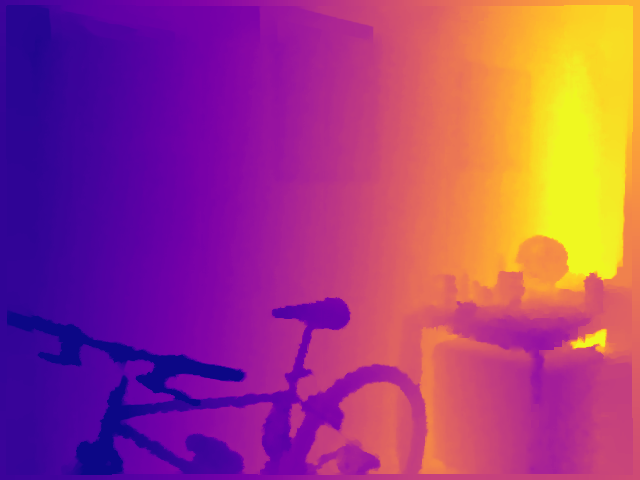}
\hfill
\includegraphics[width=0.245\textwidth,height=0.245\textheight,keepaspectratio]{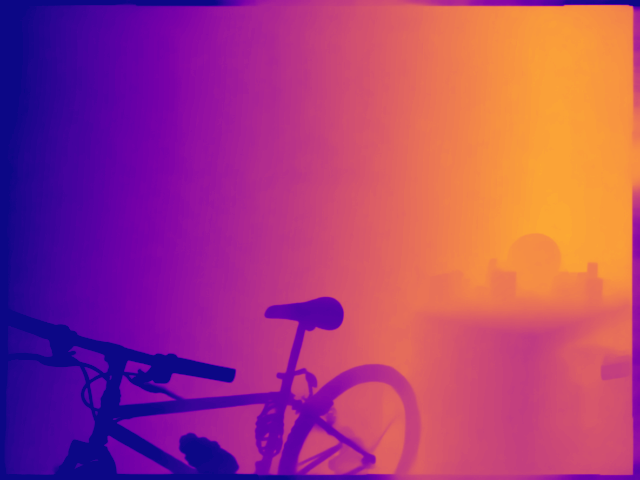}
\hfill
\includegraphics[width=0.245\textwidth,height=0.245\textheight,keepaspectratio]{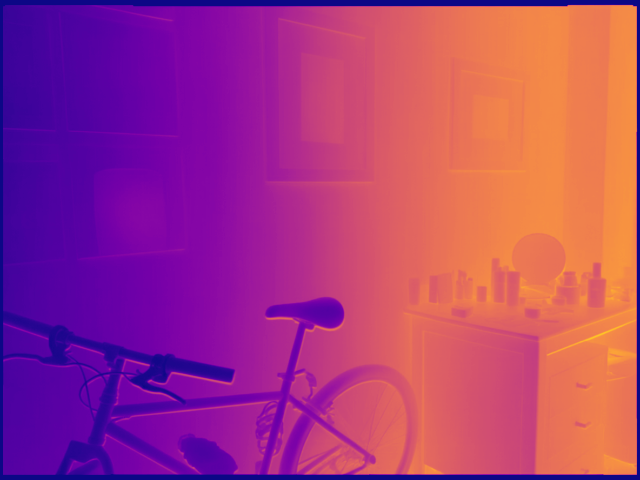}
\hfill
\includegraphics[width=0.245\textwidth,height=0.245\textheight,keepaspectratio]{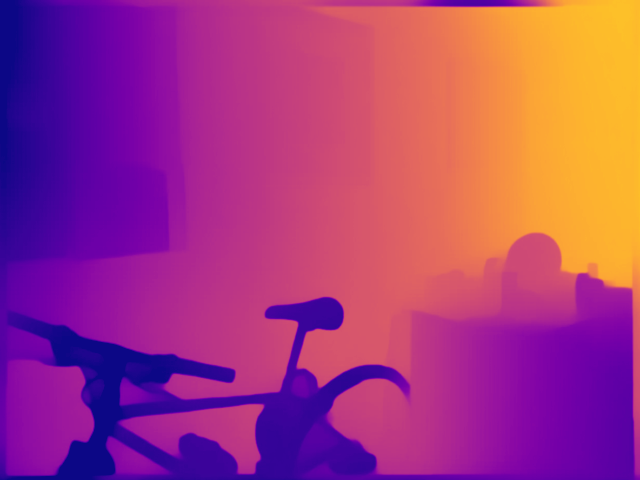}
\caption{Qualitative depth estimation on NYUv2. Top row, left to right: input image and raw grayscale depth maps generated by Qwen-Image-Edit, FireRed-Image-Edit, and LongCat-Image-Edit, where brighter pixels indicate closer surfaces. Bottom row: ground-truth depth and the corresponding affine-aligned decoded predictions, visualized with a shared false-color scale (purple = near, yellow = far).}
\label{fig:depth_qual}
\end{figure*}

\paragraph{Results.}
Figure~\ref{fig:depth_qual} shows generated and decoded depth outputs. Table~\ref{tab:depth} reports quantitative results.

\begin{table}[t]
\centering
\small
\begin{tabular}{llccc}
\toprule
Dataset & Model & $\delta_1 \uparrow$ & AbsRel$\downarrow$ & RMSE$\downarrow$ \\
\midrule
NYUv2 & Vision Banana~\cite{gabeur2026image}$^\dagger$ & 0.948 & 0.081 & -- \\
\midrule
NYUv2 & Qwen-Image-Edit & 0.756 & 0.176 & 0.585 \\
NYUv2 & FireRed-Image-Edit & 0.597 & 0.270 & 0.810 \\
NYUv2 & LongCat-Image-Edit & 0.822 & 0.141 & 0.485 \\
\midrule
DIODE In. & Qwen-Image-Edit & 0.868 & 0.123 & 0.452 \\
DIODE In. & FireRed-Image-Edit & 0.809 & 0.162 & 0.577 \\
DIODE In. & LongCat-Image-Edit & 0.833 & 0.145 & 0.515 \\
\midrule
DIODE Out. & Qwen-Image-Edit & 0.529 & 0.578 & 7.20 \\
DIODE Out. & FireRed-Image-Edit & 0.479 & 0.642 & 7.83 \\
DIODE Out. & LongCat-Image-Edit & 0.548 & 0.539 & 6.84 \\
\bottomrule
\end{tabular}
\caption{Depth estimation results. Zero-shot editors are evaluated with affine alignment to measure relative scene geometry. $^\dagger$Vision Banana is instruction-tuned and reports raw metric depth, so its numbers are shown only as a reference and are not directly comparable to our affine-aligned results.}
\label{tab:depth}
\end{table}

\paragraph{Analysis.}
All three models recover meaningful relative depth structure despite receiving no depth supervision. LongCat-Image-Edit performs best on NYUv2 ($\delta_1{=}0.822$) and DIODE Outdoor ($\delta_1{=}0.548$), while Qwen-Image-Edit leads on DIODE Indoor ($\delta_1{=}0.868$). Performance drops substantially on DIODE Outdoor (best $\delta_1{=}0.548$), suggesting that current zero-shot editors handle bounded indoor geometry more reliably than large-scale outdoor depth. FireRed-Image-Edit consistently ranks third but still achieves strong results on DIODE Indoor ($\delta_1{=}0.809$), indicating that the depth signal is not tied to a single model or dataset.

\subsection{Surface Normal Estimation}

\paragraph{Experimental Setup.}
Surface normal estimation tests whether an editor can infer local 3D orientation. We evaluate on NYUv2, using 654 validation images with ground-truth normals from~\citet{ladicky2014pulling}.

\paragraph{Implementation.}
Each model is prompted to generate a standard RGB normal map (e.g.\ ``Generate a surface normal estimation visualization of this image. Use the standard normal map color convention: surfaces facing left are pinkish-red, surfaces facing up are light green, surfaces facing the camera are light blue/purple.''). We resize generated outputs with bilinear interpolation and decode them into unit normal vectors. Axis conventions are calibrated per model--dataset pair using the procedure in Section~\ref{sec:method}.

\paragraph{Metrics.}
We report mean and median angular error in degrees. We also report A11, A22, and A30: the percentage of pixels with angular error below $11.25^\circ$, $22.5^\circ$, and $30^\circ$.

\begin{figure*}[t]
\centering
\includegraphics[width=0.245\textwidth,height=0.245\textheight,keepaspectratio]{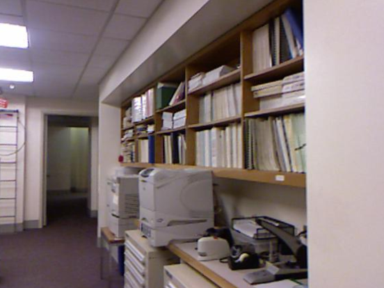}
\hfill
\includegraphics[width=0.245\textwidth,height=0.245\textheight,keepaspectratio]{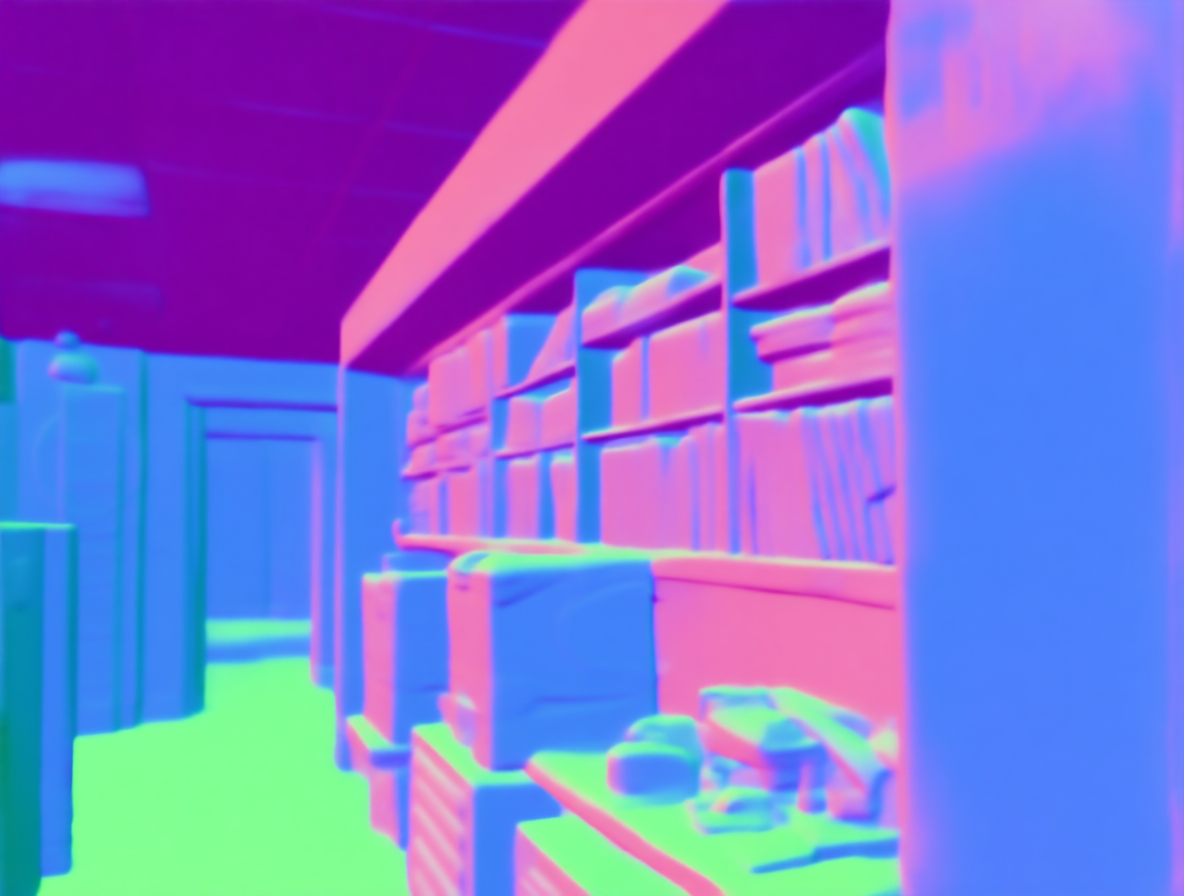}
\hfill
\includegraphics[width=0.245\textwidth,height=0.245\textheight,keepaspectratio]{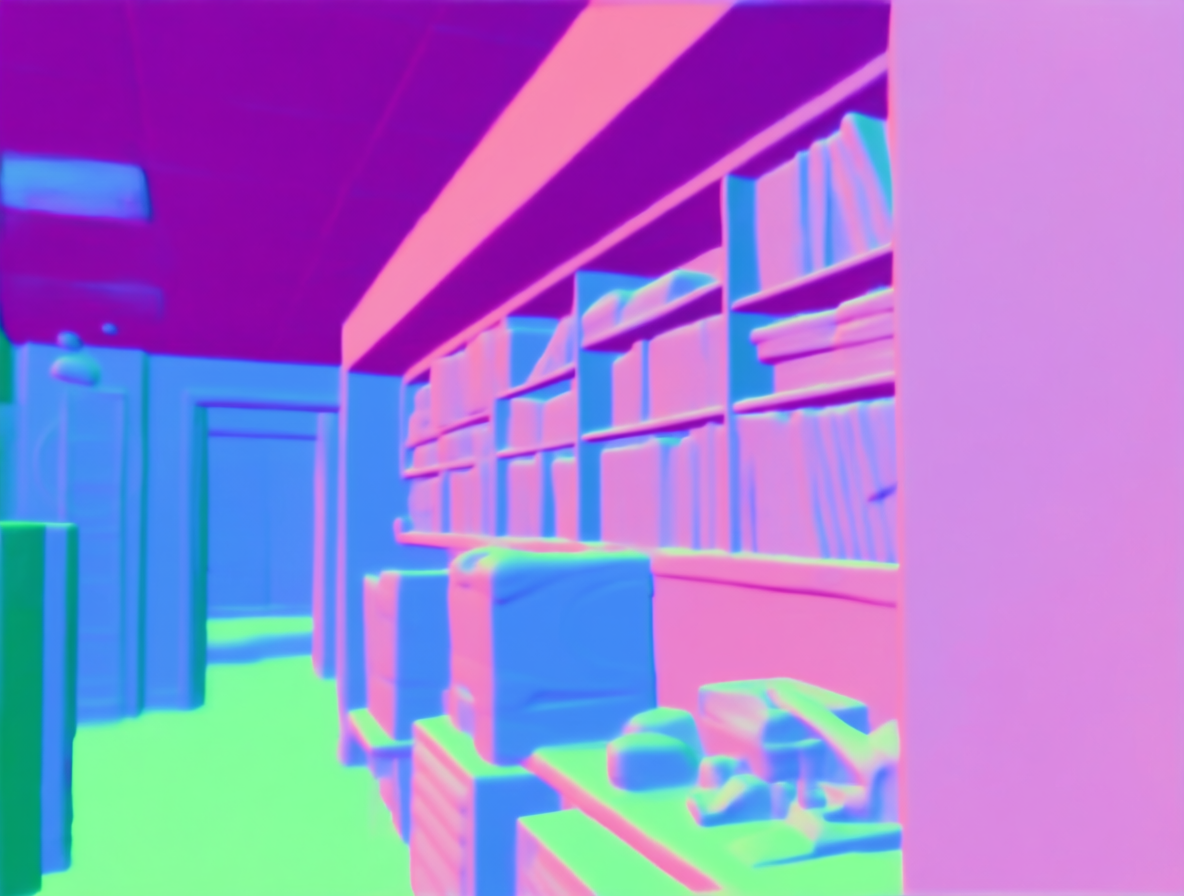}
\hfill
\includegraphics[width=0.245\textwidth,height=0.245\textheight,keepaspectratio]{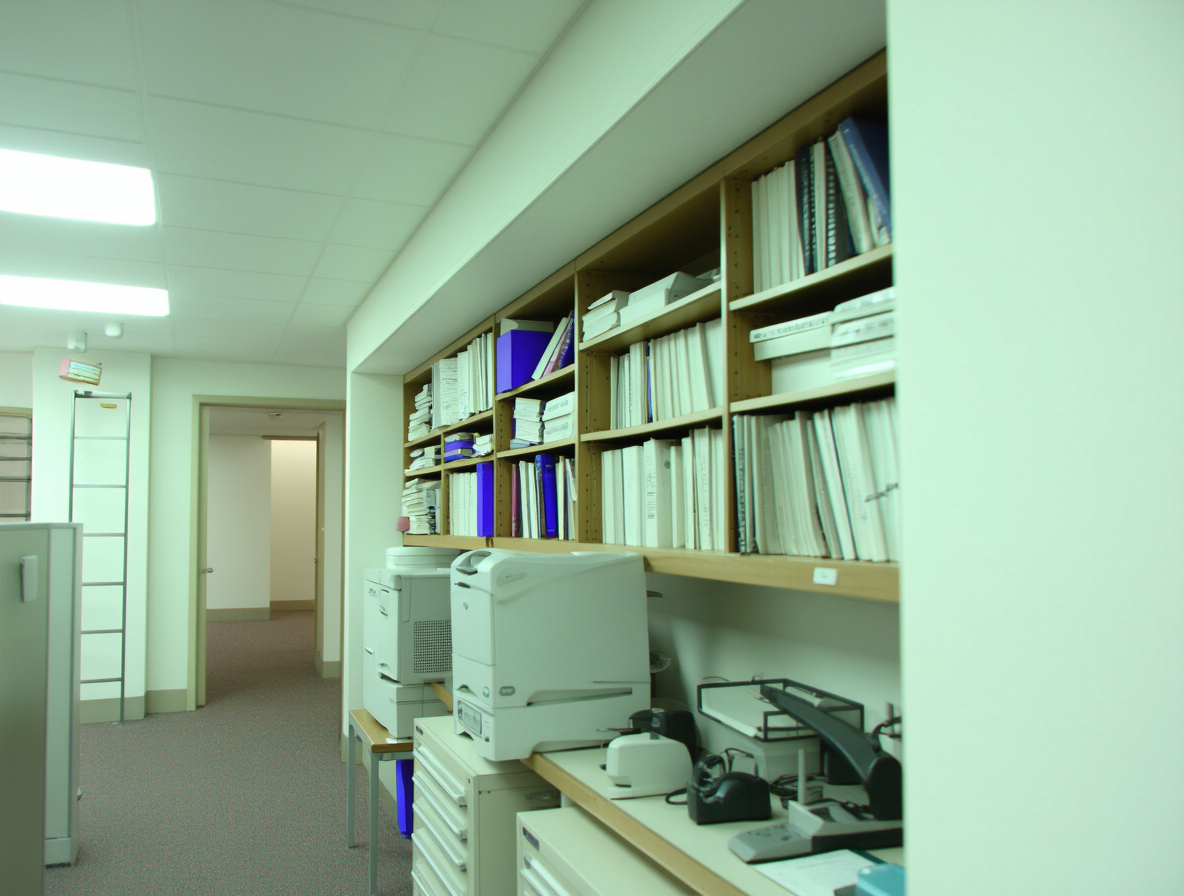}

\vspace{0.2em}

\includegraphics[width=0.245\textwidth,height=0.245\textheight,keepaspectratio]{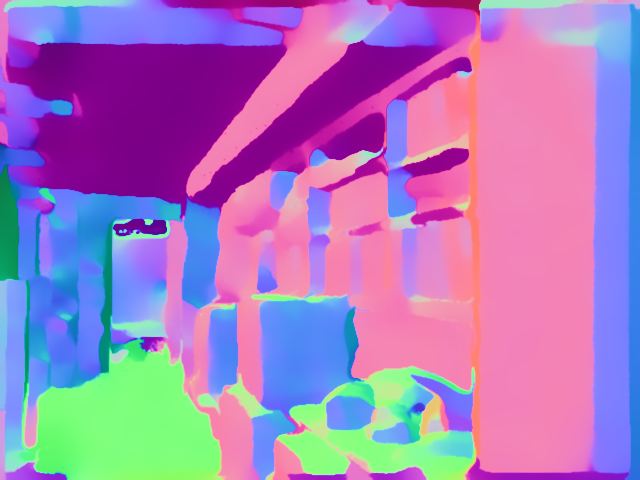}
\hfill
\includegraphics[width=0.245\textwidth,height=0.245\textheight,keepaspectratio]{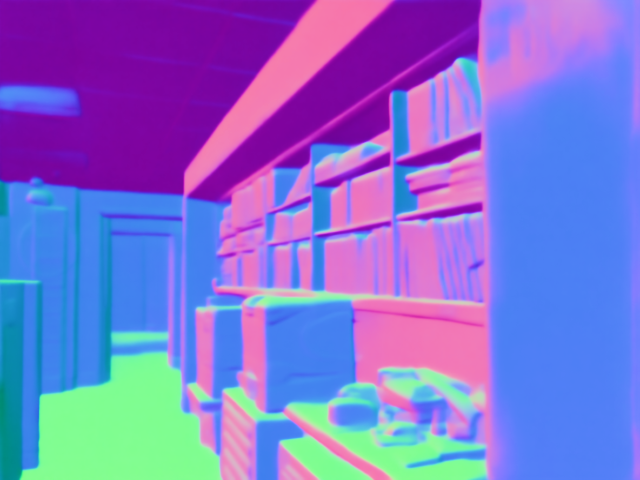}
\hfill
\includegraphics[width=0.245\textwidth,height=0.245\textheight,keepaspectratio]{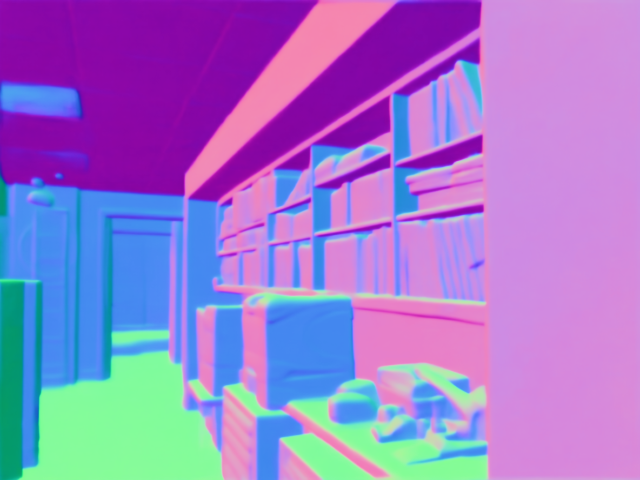}
\hfill
\includegraphics[width=0.245\textwidth,height=0.245\textheight,keepaspectratio]{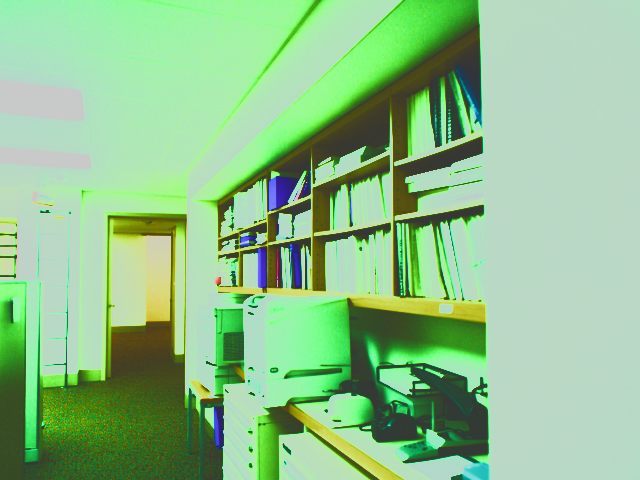}
\caption{Qualitative surface normal estimation on NYUv2. Top row, left to right: input image and raw normal maps generated by Qwen-Image-Edit, FireRed-Image-Edit, and LongCat-Image-Edit. Bottom row: ground-truth normal map and the corresponding decoded predictions after axis calibration. Colors encode surface orientation using the normal-map convention, with RGB channels corresponding to the normalized $(x,y,z)$ normal components.}
\label{fig:normals_qual}
\end{figure*}

\paragraph{Results.}
Figure~\ref{fig:normals_qual} shows generated and decoded normal maps, and Table~\ref{tab:normals} reports quantitative results.

\begin{table}[t]
\centering
\small
\begin{tabular}{lccccc}
\toprule
Model & Mean$\downarrow$ & Med.$\downarrow$ & A11$\uparrow$ & A22$\uparrow$ & A30$\uparrow$ \\
\midrule
Vision Banana~\cite{gabeur2026image} & 17.78 & 8.88 & -- & -- & -- \\
Marigold~\cite{ke2024marigold} & 20.86 & 11.13 & -- & -- & -- \\
\midrule
Qwen-Image-Edit & 20.51 & 13.93 & 43.2 & 69.7 & 78.7 \\
FireRed-Image-Edit & 17.69 & 11.10 & 51.6 & 76.4 & 83.4 \\
LongCat-Image-Edit & 67.86 & 64.29 & 3.2 & 10.2 & 15.9 \\
\bottomrule
\end{tabular}
\caption{Surface normal estimation on NYUv2. Mean and median are angular errors in degrees. A11/A22/A30 denote the percentage of pixels with angular error below $11.25^\circ$, $22.5^\circ$, and $30^\circ$. Vision Banana and Marigold are tuned references.}
\label{tab:normals}
\end{table}

\paragraph{Analysis.}
FireRed-Image-Edit achieves the best zero-shot result ($17.69^\circ$ mean error), comparable to the instruction-tuned Vision Banana ($17.78^\circ$) without any task-specific training. Qwen-Image-Edit ($20.51^\circ$) is close behind, and both models outperform the fine-tuned Marigold reference ($20.86^\circ$). LongCat-Image-Edit is an outlier at $67.86^\circ$; qualitative inspection shows that it often generates stylized images rather than standard normal maps. This task therefore measures both local surface understanding and adherence to the normal-map convention. The strong performance of FireRed and Qwen suggests that zero-shot normal estimation is not limited to a single editor, while LongCat highlights the importance of output-format following.

\subsection{Semantic Segmentation}

\paragraph{Experimental Setup.}
Semantic segmentation tests whether an editor can localize semantic categories through color-coded mask generation. We evaluate on the 500-image Cityscapes validation set~\citep{cordts2016cityscapes}.

\paragraph{Implementation.}
For each image, we construct the prompt from an oracle image-level class list: only classes present in the ground-truth annotation are listed. In the 19-class evaluation, each class is assigned an intuitive color name and the prompt takes a simple color-mapping form (e.g.\ ``Convert this photo into a color-coded map: the road red, the car white, \ldots and everything else black.''). In the 7-category coarse evaluation, sub-classes are first grouped into super-categories (flat, construction, nature, sky, human, vehicle, object), and the prompt explicitly requests flat rendering (e.g.\ ``Turn this image into a flat segmentation mask using only solid colors. Paint all roads and sidewalks solid red, \ldots and everything else solid black. No textures, no gradients.''). Generated outputs are resized with nearest-neighbor interpolation and decoded by assigning each pixel to the nearest prompted color. This oracle class-list setting evaluates mask generation and palette following rather than open-vocabulary class discovery.

\paragraph{Metrics.}
Given the oracle image-level class list in the prompt, we report mIoU and pixel accuracy in the Cityscapes label space.

\begin{figure*}[t]
\centering
\includegraphics[width=0.245\textwidth,height=0.245\textheight,keepaspectratio]{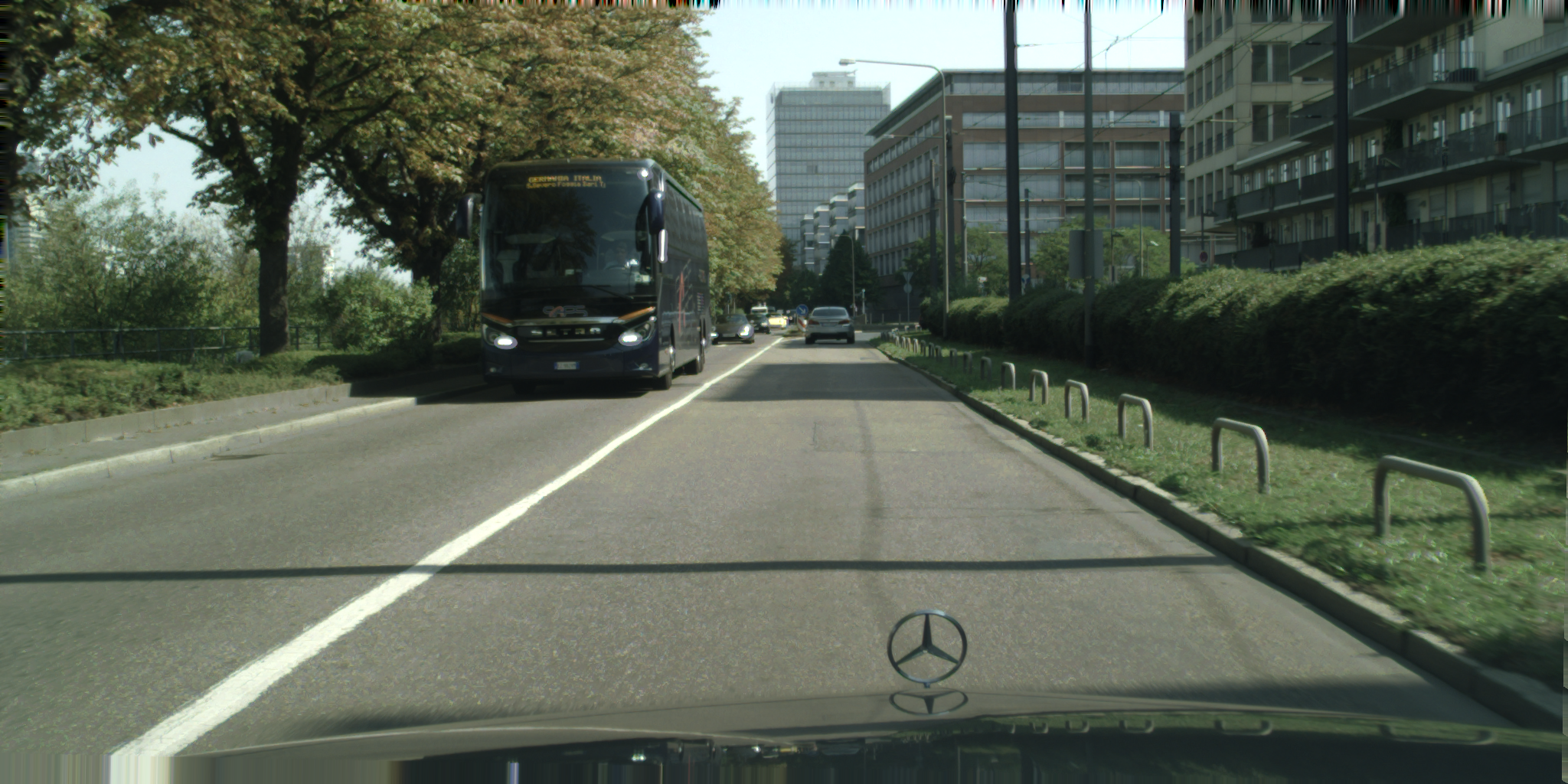}
\hfill
\includegraphics[width=0.245\textwidth,height=0.245\textheight,keepaspectratio]{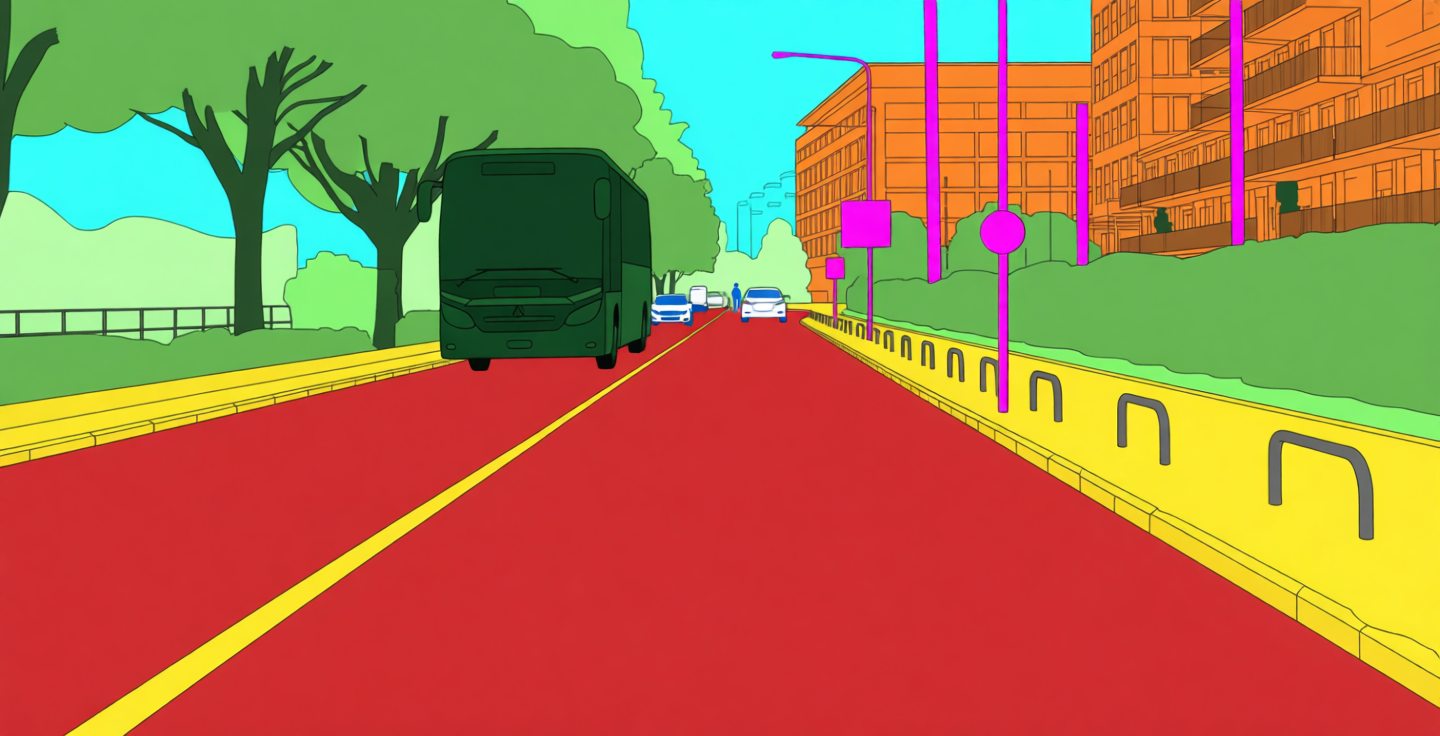}
\hfill
\includegraphics[width=0.245\textwidth,height=0.245\textheight,keepaspectratio]{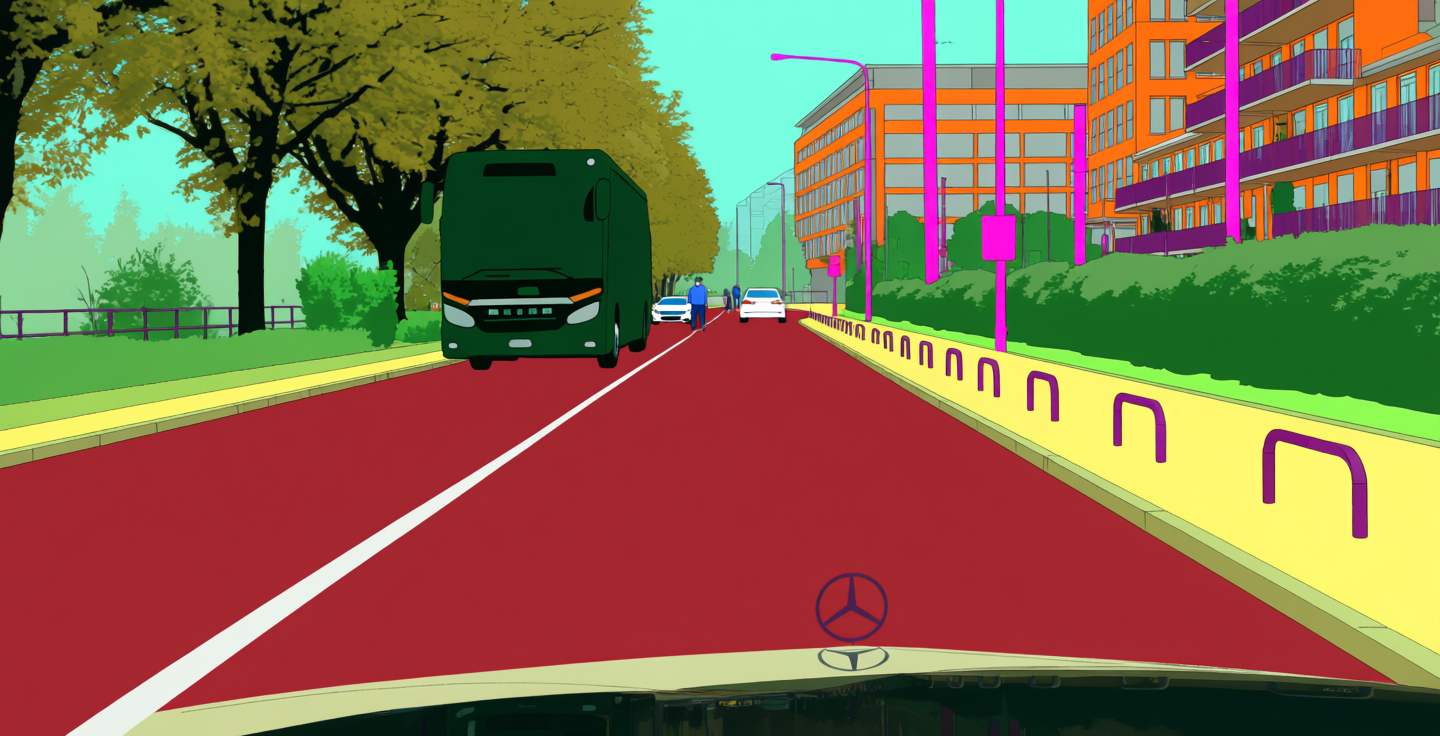}
\hfill
\includegraphics[width=0.245\textwidth,height=0.245\textheight,keepaspectratio]{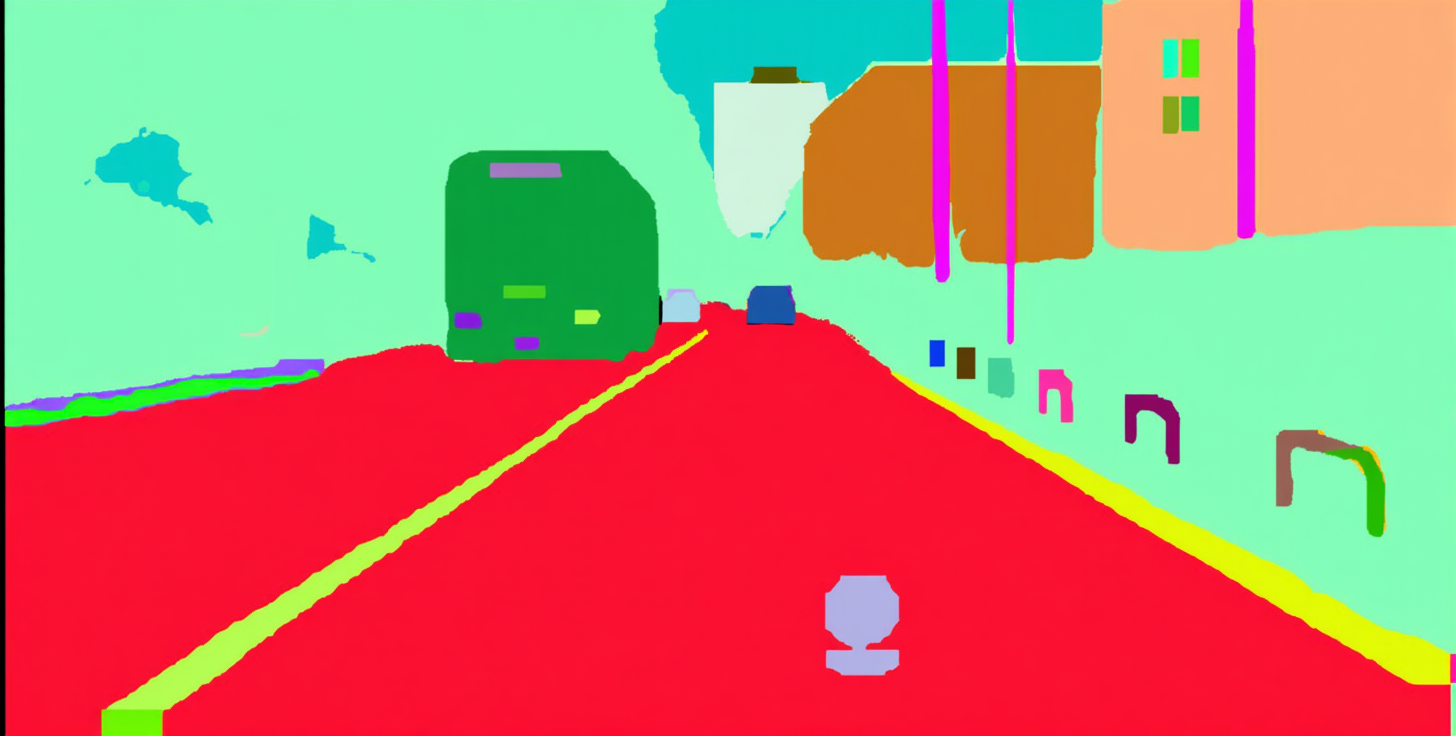}

\vspace{0.2em}

\includegraphics[width=0.245\textwidth,height=0.245\textheight,keepaspectratio]{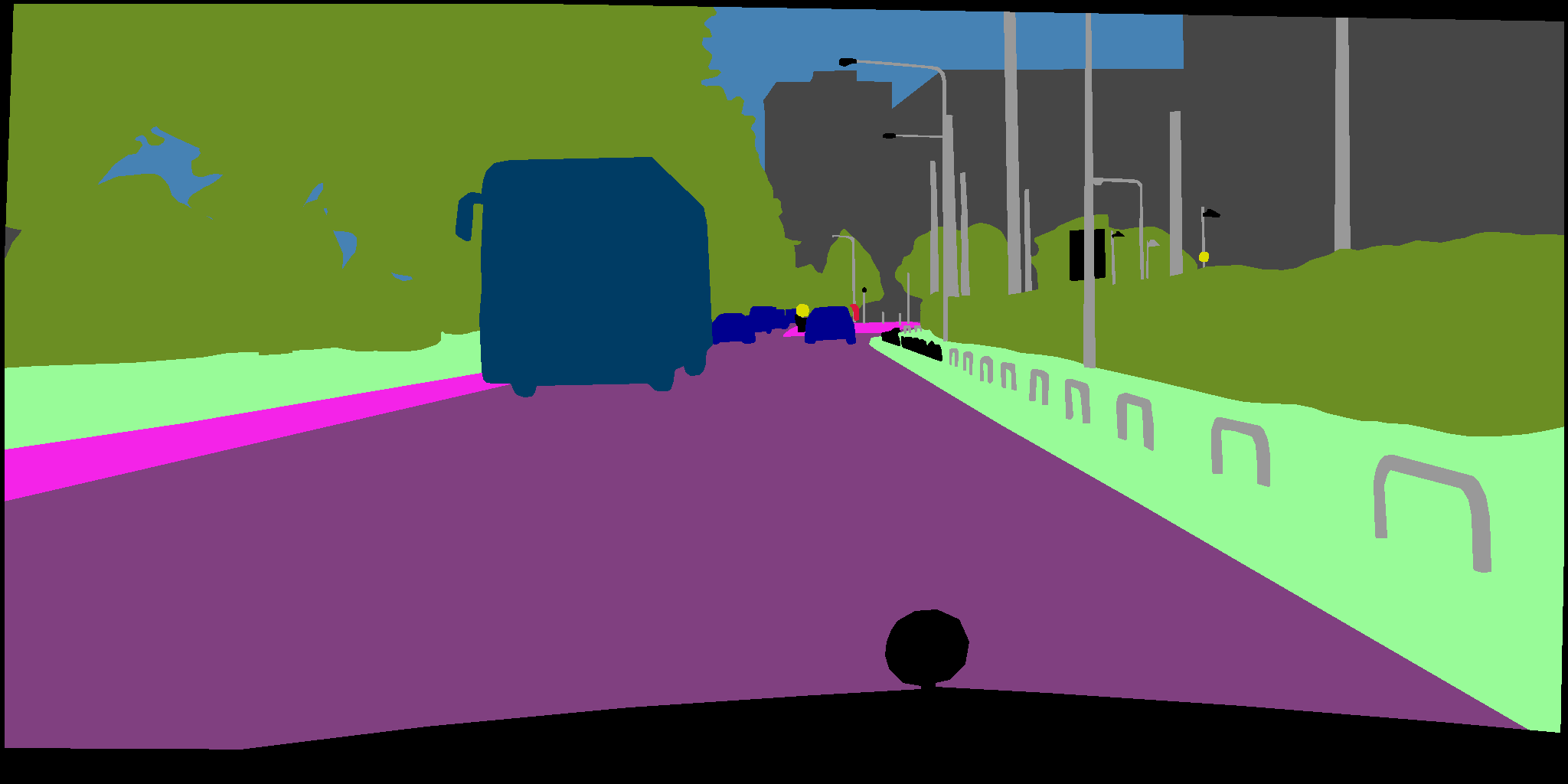}
\hfill
\includegraphics[width=0.245\textwidth,height=0.245\textheight,keepaspectratio]{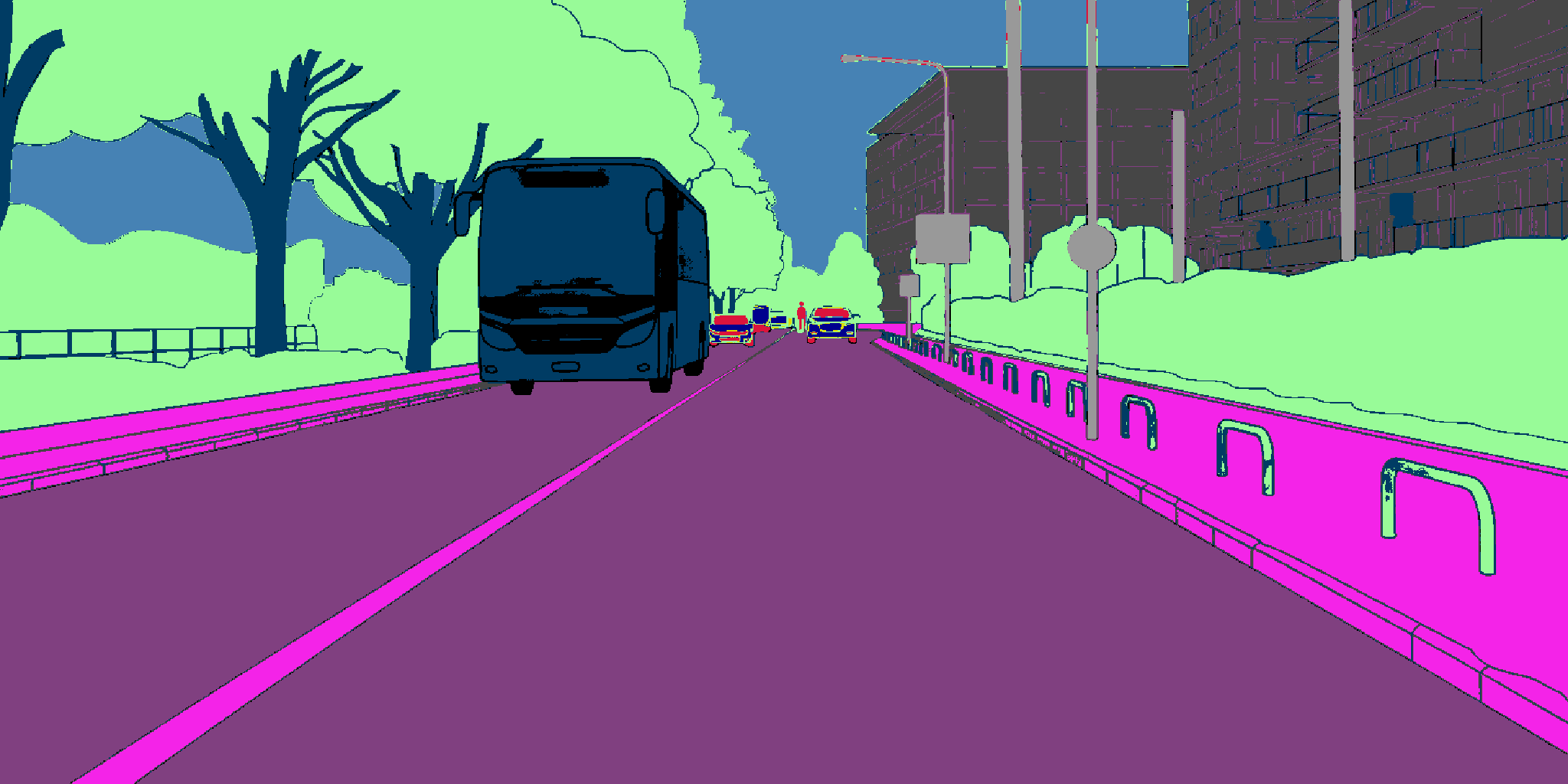}
\hfill
\includegraphics[width=0.245\textwidth,height=0.245\textheight,keepaspectratio]{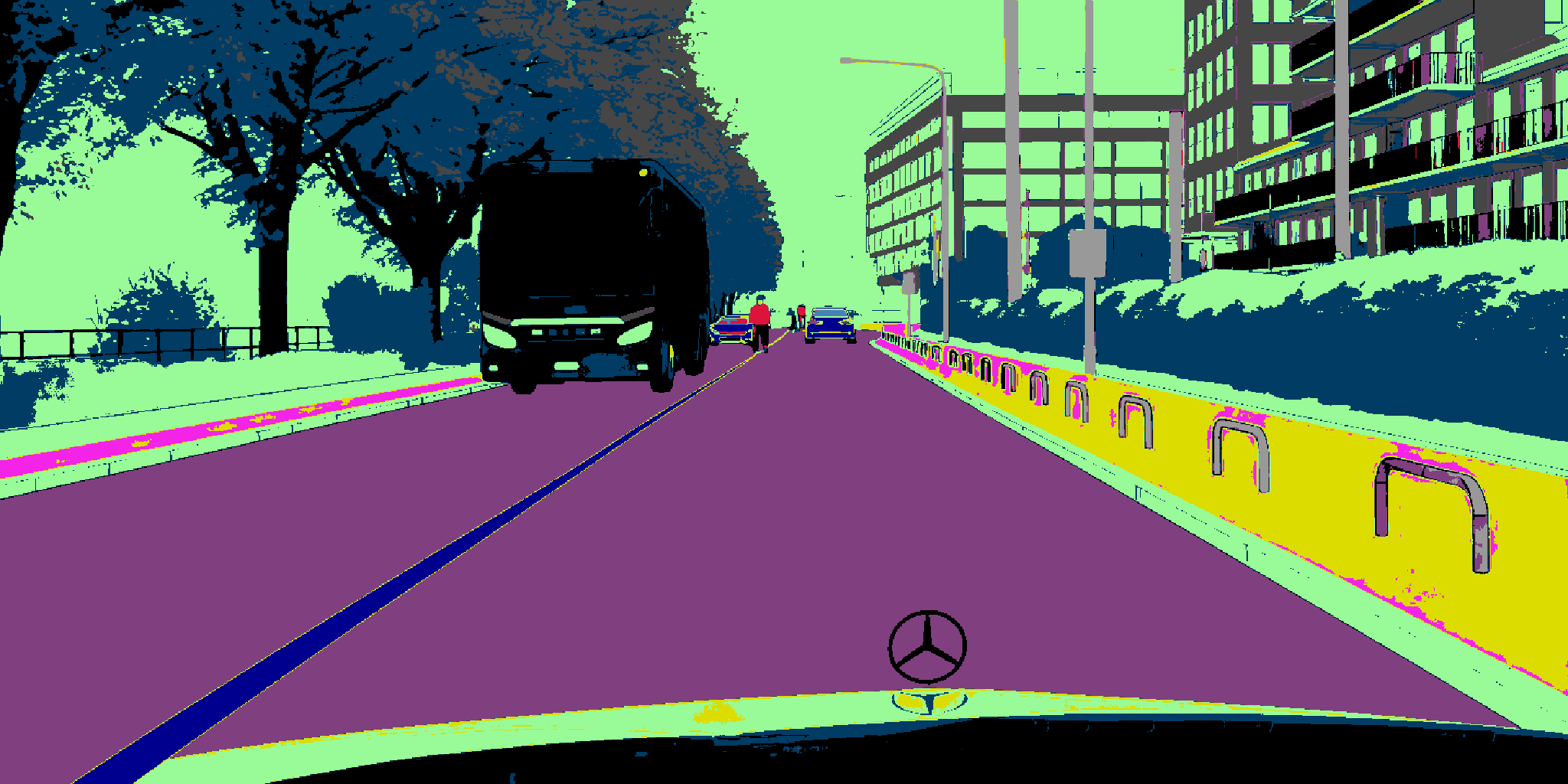}
\hfill
\includegraphics[width=0.245\textwidth,height=0.245\textheight,keepaspectratio]{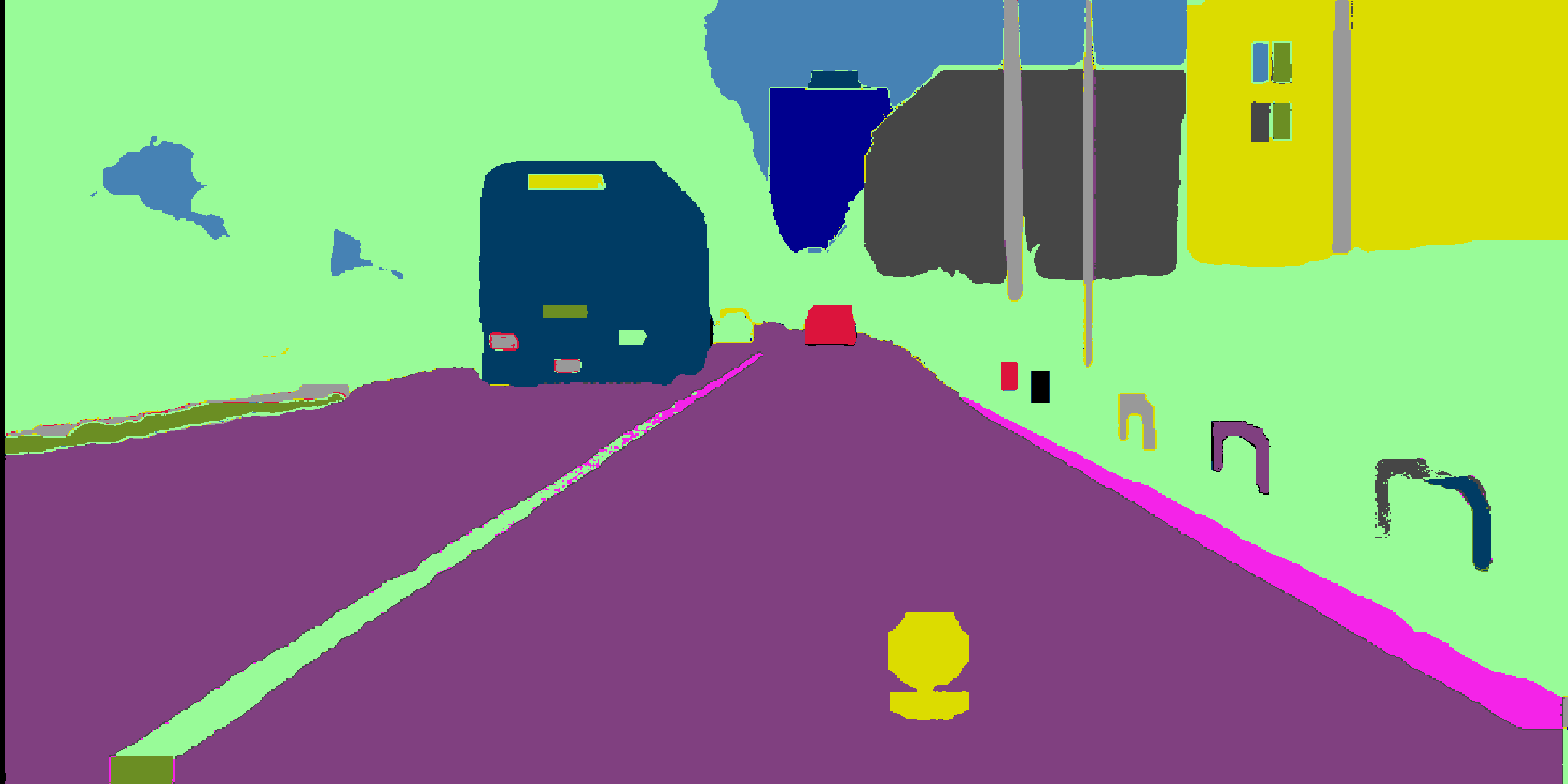}
\caption{Qualitative semantic segmentation on Cityscapes (\textbf{19-class}). Top row, left to right: input image and raw color masks generated by Qwen-Image-Edit, FireRed-Image-Edit, and LongCat-Image-Edit from a natural-language segmentation prompt. Bottom row: ground-truth label map (left) and the corresponding decoded predictions from each model. The prompt specifies the target classes and their colors; decoded masks are obtained by assigning each pixel to its nearest prompt-specified color.}
\label{fig:semantic_qual}
\end{figure*}

\begin{figure*}[t]
\centering
\includegraphics[width=0.245\textwidth,height=0.245\textheight,keepaspectratio]{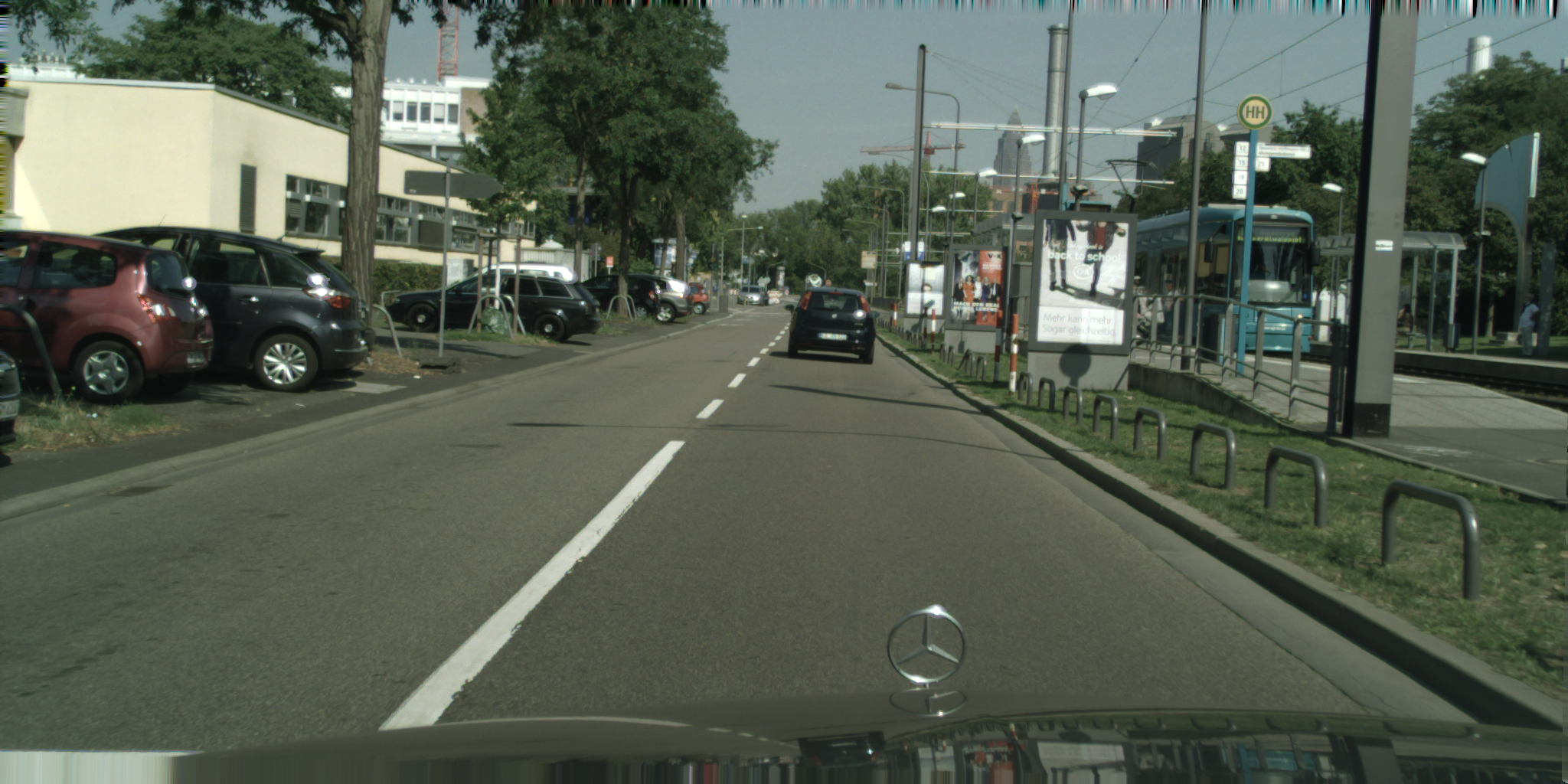}
\hfill
\includegraphics[width=0.245\textwidth,height=0.245\textheight,keepaspectratio]{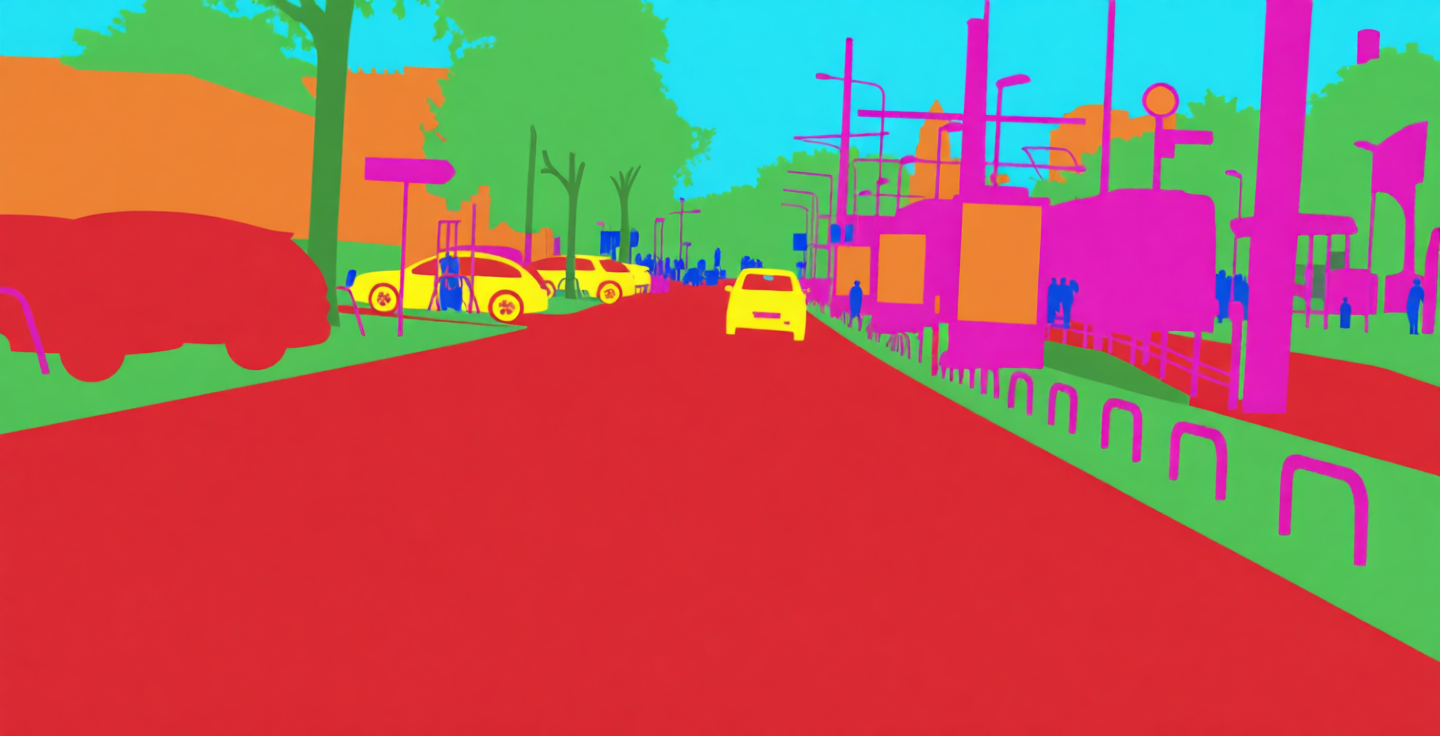}
\hfill
\includegraphics[width=0.245\textwidth,height=0.245\textheight,keepaspectratio]{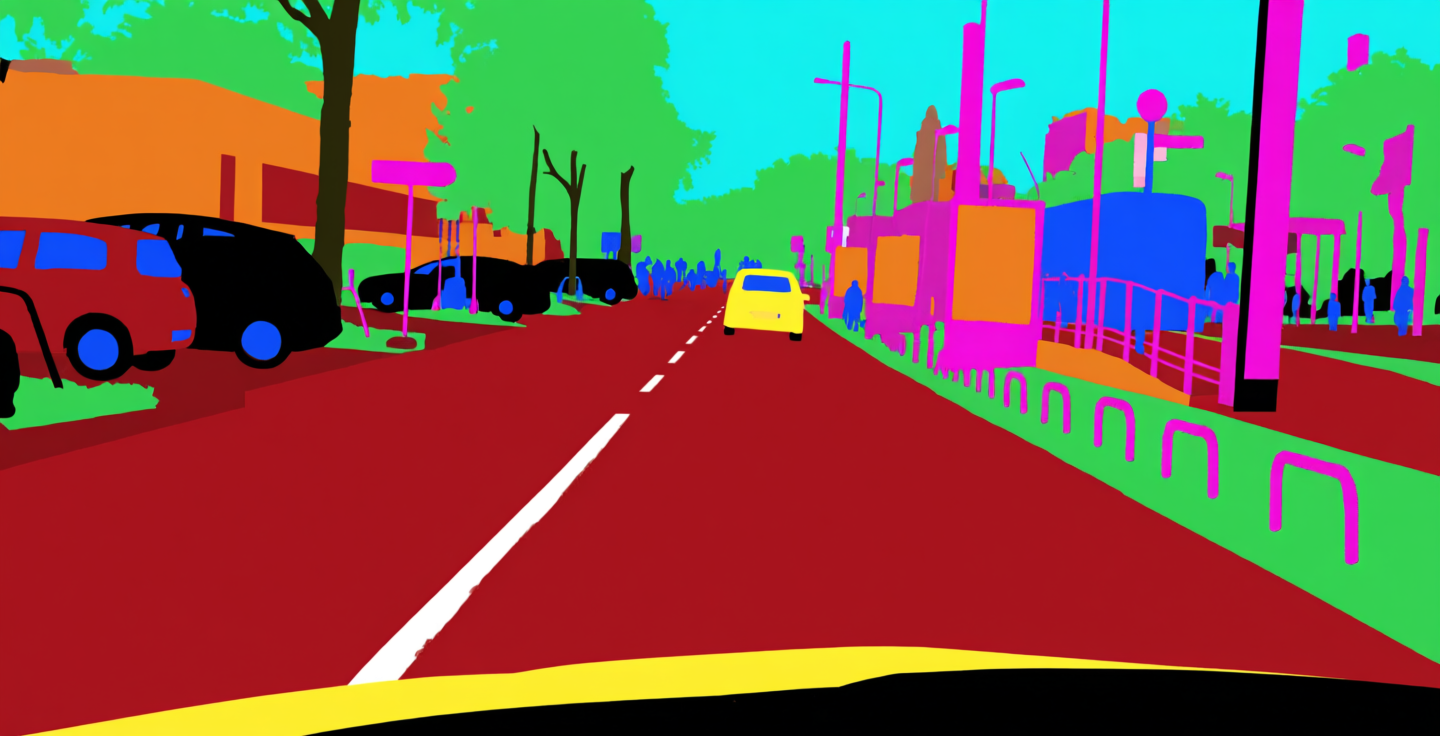}
\hfill
\includegraphics[width=0.245\textwidth,height=0.245\textheight,keepaspectratio]{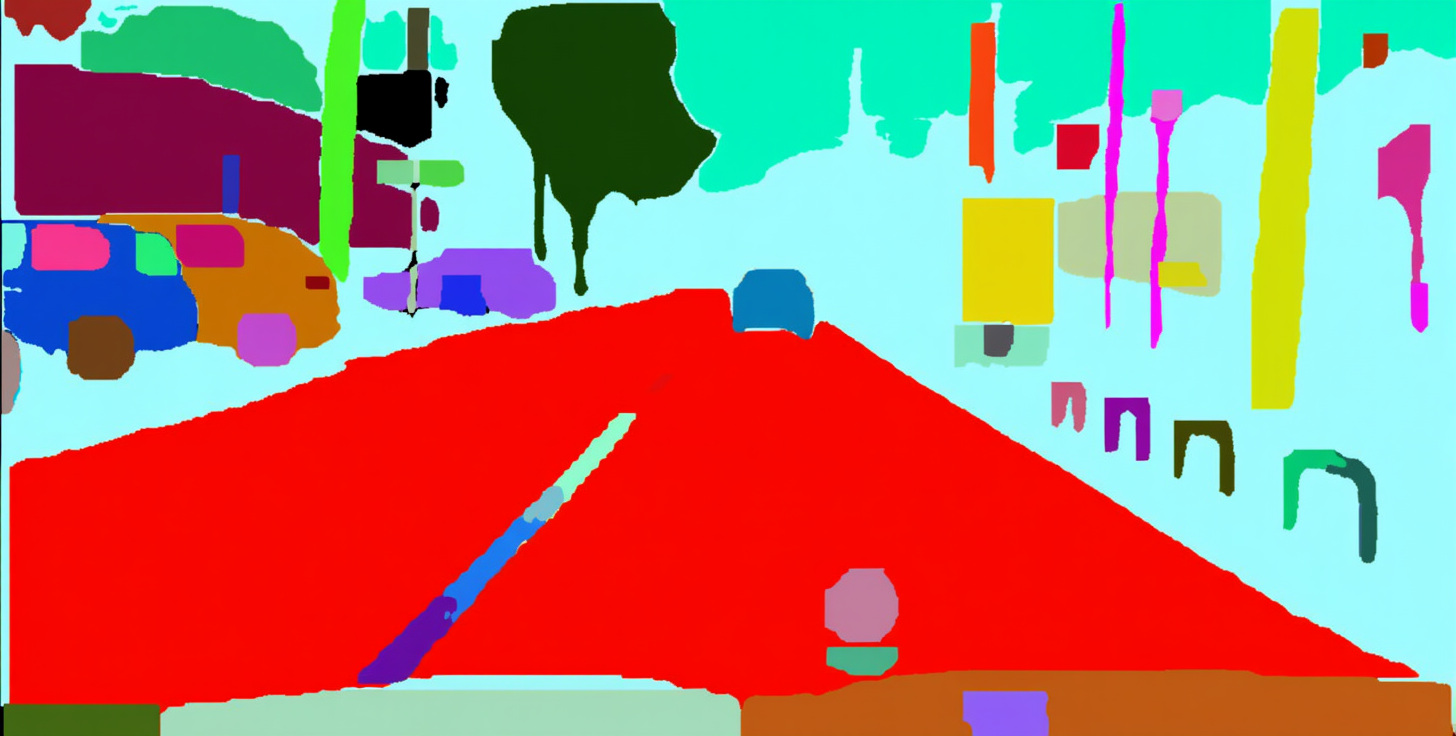}

\vspace{0.2em}

\includegraphics[width=0.245\textwidth,height=0.245\textheight,keepaspectratio]{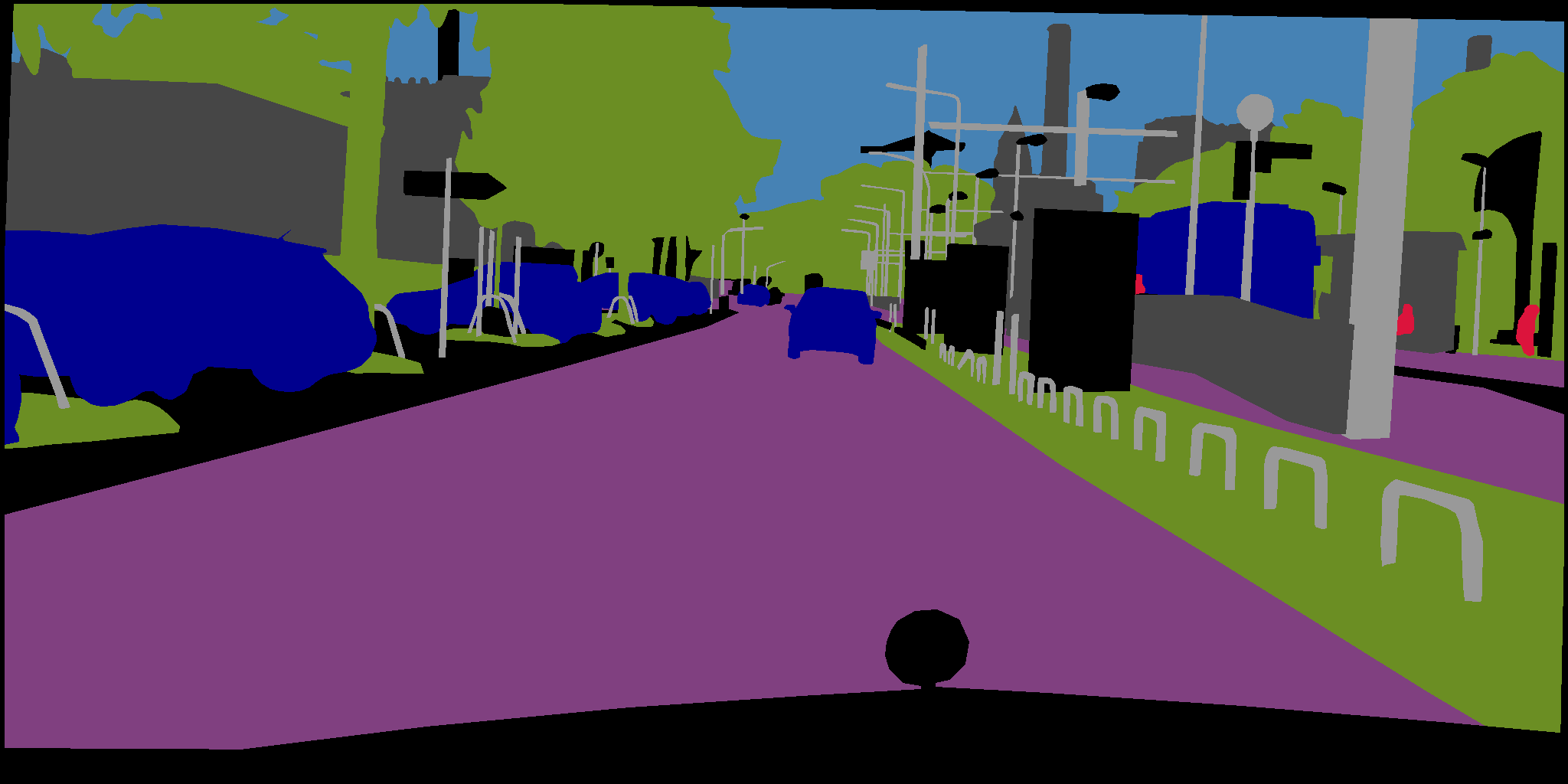}
\hfill
\includegraphics[width=0.245\textwidth,height=0.245\textheight,keepaspectratio]{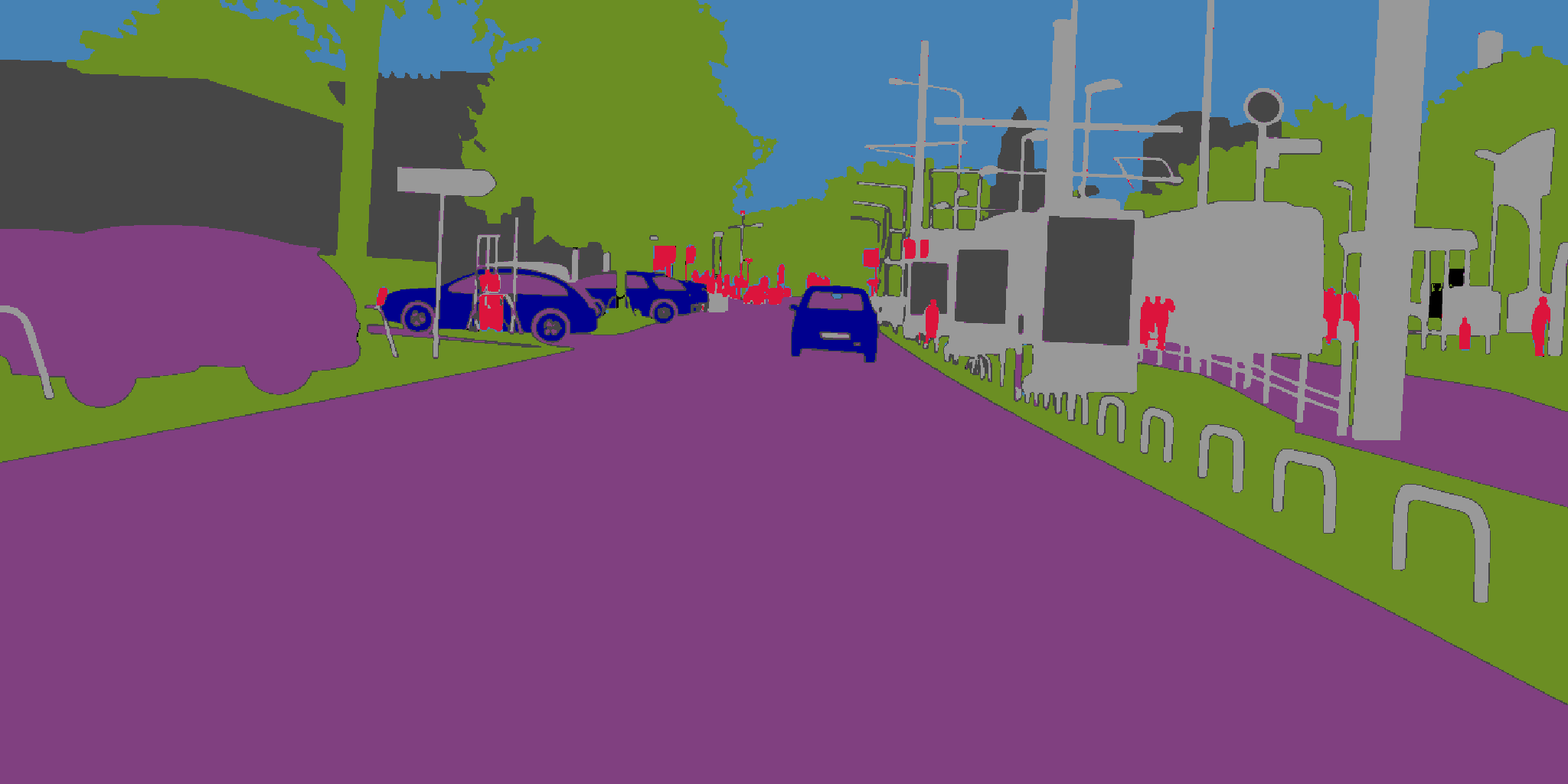}
\hfill
\includegraphics[width=0.245\textwidth,height=0.245\textheight,keepaspectratio]{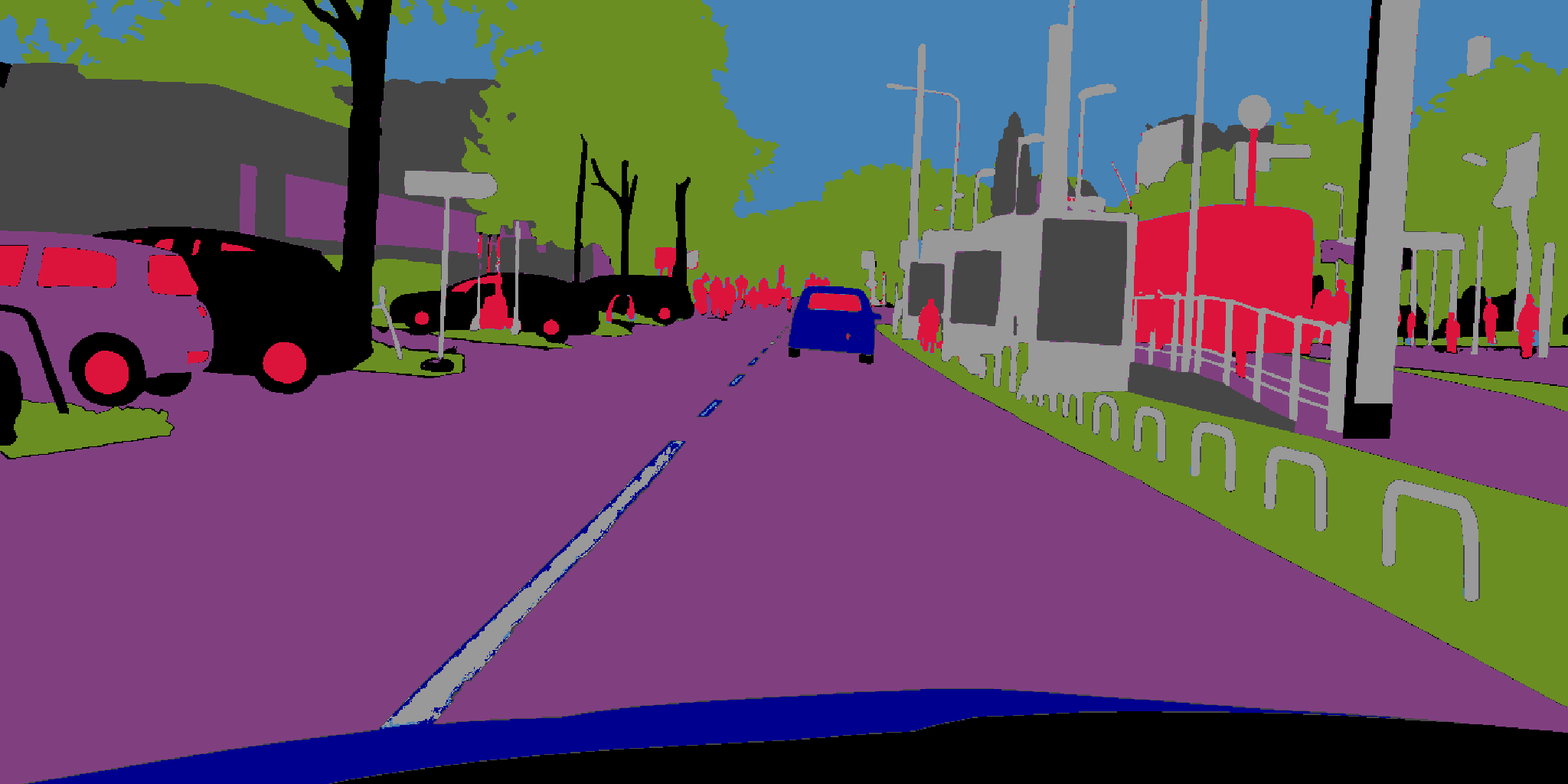}
\hfill
\includegraphics[width=0.245\textwidth,height=0.245\textheight,keepaspectratio]{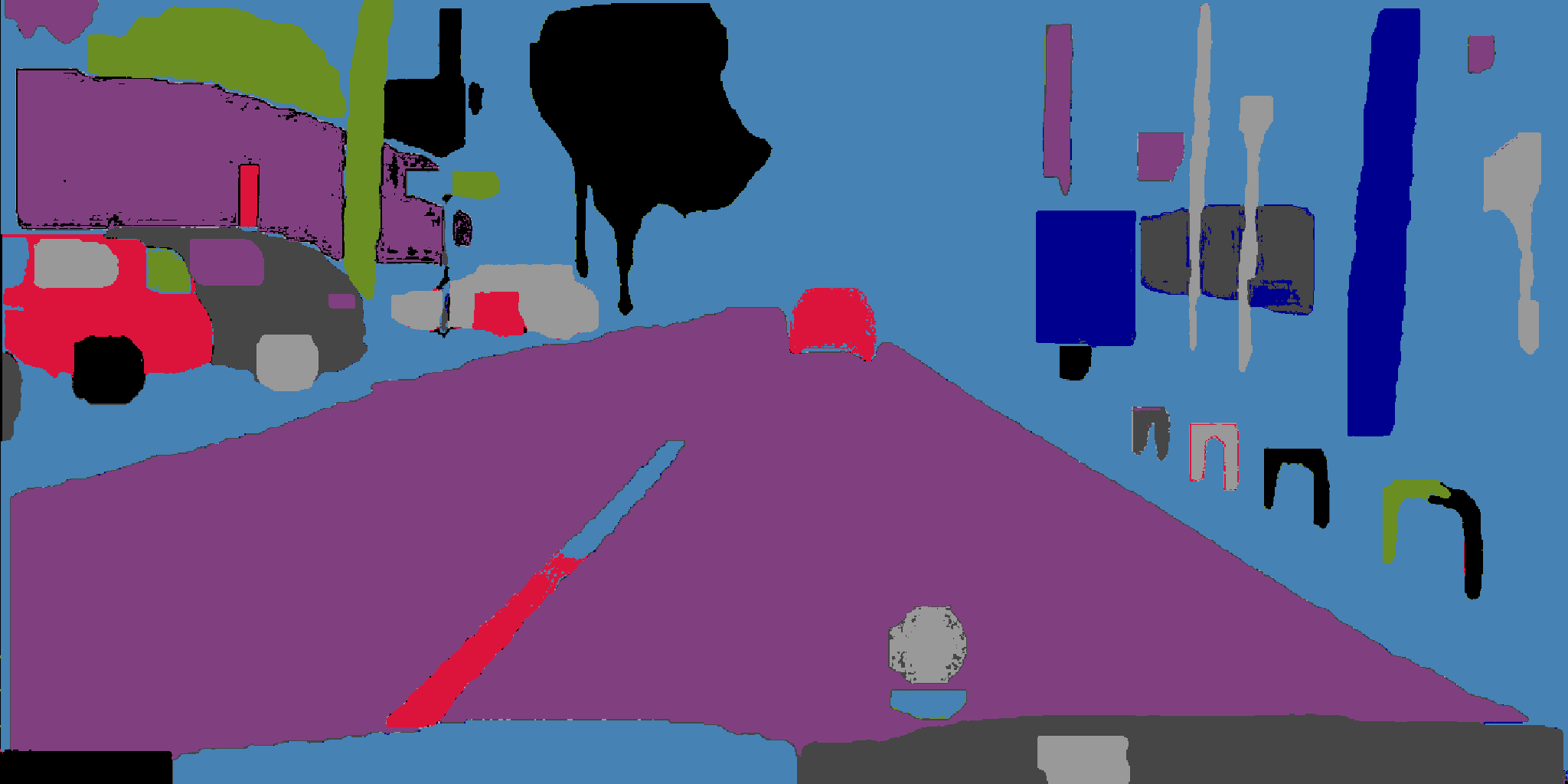}
\caption{Qualitative semantic segmentation on Cityscapes (\textbf{7-category}). Top row, left to right: input image and raw color masks generated by Qwen-Image-Edit, FireRed-Image-Edit, and LongCat-Image-Edit from a natural-language segmentation prompt. Bottom row: ground-truth label map (left) and the corresponding decoded predictions from each model. The prompt specifies the target classes and their colors; decoded masks are obtained by assigning each pixel to its nearest prompt-specified color.}
\label{fig:semantic_qual_7}
\end{figure*}

\paragraph{Results.}
Figures~\ref{fig:semantic_qual} and~\ref{fig:semantic_qual_7} show qualitative segmentation outputs for the 19-class and 7-category settings. Table~\ref{tab:semantic} reports quantitative results.

\begin{table}[t]
\centering
\small
\begin{tabular}{llcc}
\toprule
Granularity & Model & mIoU $\uparrow$ & Pixel Acc. $\uparrow$ \\
\midrule
19-class & Vision Banana~\cite{gabeur2026image} & 69.9 & -- \\
\midrule
19-class & Qwen-Image-Edit & 25.7 & 55.4 \\
19-class & FireRed-Image-Edit & 20.5 & 35.6 \\
19-class & LongCat-Image-Edit & 17.1 & 48.1 \\
\midrule
7-category & Qwen-Image-Edit & 49.5 & 69.4 \\
7-category & FireRed-Image-Edit & 44.5 & 61.6 \\
7-category & LongCat-Image-Edit & 31.4 & 60.9 \\
\bottomrule
\end{tabular}
\caption{Semantic segmentation on Cityscapes (500 val images). Our zero-shot editors are evaluated with oracle image-level class lists. Vision Banana is instruction-tuned and shown only as a reference.}
\label{tab:semantic}
\end{table}

\paragraph{Analysis.}
Semantic segmentation is a stricter test than depth or normals because the model must both identify semantic regions and render them with the requested colors. Under the oracle class-list setting, Qwen-Image-Edit achieves 25.7 mIoU at the 19-class level. This is far below the instruction-tuned Vision Banana (69.9), but it shows that the model can produce non-trivial dense masks without segmentation training. When classes are grouped into 7 coarse categories, mIoU rises to 49.5 for Qwen and 44.5 for FireRed, indicating that the models capture broad scene layout (road vs.\ building vs.\ vegetation vs.\ sky) more reliably than fine-grained Cityscapes categories.

Qualitative inspection suggests that semantic errors come from two sources. First, a model may localize an object but render it with the wrong requested color; such outputs show visual understanding but fail under color-based decoding. Second, as the number of classes grows, the RGB-mask interface becomes harder: the prompt must assign many distinguishable colors, including less common color names, and the editor must reproduce them as flat, solid regions rather than natural image colors. This makes semantic segmentation particularly sensitive to instruction following and palette execution. The gap to Vision Banana therefore reflects both segmentation quality and the absence of instruction tuning for dense color-mask formats.

\subsection{Cross-Model Comparison}

A central question of this study is whether zero-shot vision ability is shared across independently trained image-editing models or concentrated in a single model. The results show both shared ability and task-dependent differences. For depth estimation, all three editors recover relative geometry, with LongCat leading on NYUv2 and DIODE Outdoor while Qwen leads on DIODE Indoor; FireRed consistently ranks third. For surface normals, FireRed and Qwen perform strongly, while LongCat struggles to follow the normal-map convention. For semantic segmentation, Qwen is strongest at the 19-class level, while Qwen and FireRed are close under the 7-category evaluation.

These patterns suggest that image-editing pretraining provides useful visual representations across models, but quantitative performance also depends on whether a model can follow the required output format. Depth is relatively forgiving because affine alignment can recover relative geometry from imperfect intensity scale. Normals and segmentation are less forgiving because they require specialized RGB conventions. This distinction is important for interpreting the results: some failures reflect missing visual understanding, while others reflect a mismatch between natural image editing behavior and the artificial visualization formats required for benchmark decoding.

\subsection{Failure Modes}

We analyze failure modes that are specific to using image editors for quantitative vision tasks. For depth, the main failure is metric miscalibration: generated depth maps may preserve relative ordering while producing unrealistic absolute depth. For normals, failures often appear as over-smoothed planar regions, stylized normal-map colors, or incorrect axis conventions before calibration. For semantic segmentation, common errors include correct localization with wrong colors, palette drift, blurred boundaries, missing thin structures, and confusion among visually similar classes. These failures highlight a key limitation of the image-generation interface: a generated image can be visually plausible and semantically meaningful while still being difficult to decode into exact dense labels.

%% file: tex/6.conclusion.tex
\section{Conclusion}

We presented a systematic zero-shot evaluation of open-source image-editing models as dense vision learners. Without any task-specific fine-tuning, Qwen-Image-Edit, FireRed-Image-Edit, and LongCat-Image-Edit were evaluated on monocular depth estimation, surface normal prediction, and semantic segmentation. Across these tasks, the results show that publicly available image editors exhibit non-trivial visual understanding: they can recover relative scene geometry, predict meaningful surface orientation, and produce coarse semantic layouts from image-editing prompts.

At the same time, our results reveal an important limitation of the image-as-output interface. Quantitative performance depends not only on whether a model understands the scene, but also on whether it follows the requested visualization format precisely. Depth estimation is relatively tolerant to imperfect intensity calibration after alignment, while surface normals and semantic segmentation require stricter adherence to specialized RGB conventions. In semantic segmentation, models may localize objects but use the wrong colors, causing failures under deterministic decoding despite meaningful visual understanding.

Overall, our findings support the view that image-editing pretraining can induce zero-shot vision abilities beyond image generation. The evidence is strongest for relative geometry and surface normals, while semantic segmentation exposes output-format following as a key bottleneck. This work provides a reproducible baseline for studying open-source generative models as general-purpose vision systems, and for developing better prompting, decoding, and instruction-tuning methods that turn latent visual understanding into reliable dense predictions.

\paragraph{Limitations.}
Our evaluation has several limitations. First, zero-shot performance is sensitive to prompt design; although we use simple prompts, more effective prompts may exist. Second, semantic segmentation is evaluated with oracle image-level class lists, so it measures mask generation and palette following rather than open-vocabulary class discovery. Third, instance segmentation is excluded due to decoding instability (Section~\ref{sec:method}), leaving instance-level reasoning to future work. Finally, we do not evaluate video models or text-to-image generators, which may exhibit different emergent abilities.

\paragraph{Future Work.}
Promising directions include (i)~extending the evaluation to additional geometric tasks such as optical flow and occlusion boundary detection, (ii)~evaluating larger or more recent image-editing models as they become available, (iii)~developing more robust decoding methods for instance segmentation, and (iv)~investigating which pretraining factors (data, architecture, scale) most contribute to emergent vision abilities.